\documentclass{article}

\usepackage[final]{neurips_2025}

\usepackage[utf8]{inputenc} 
\usepackage[T1]{fontenc}    
\usepackage{hyperref}       
\usepackage{url}            
\usepackage{booktabs}       
\usepackage{amsfonts}       
\usepackage{nicefrac}       
\usepackage{microtype}      
\usepackage{xcolor}         
\usepackage{amsmath}
\usepackage{amssymb}
\usepackage{colortbl}
\usepackage{graphicx}
\usepackage{multirow}
\usepackage{float}
\usepackage{subcaption}
\usepackage{caption}
\usepackage{bbm}
\usepackage{comment}
\usepackage{wrapfig}
\usepackage{enumitem}

\graphicspath{{./images/}}

\newcommand{\CR}[1]{{#1}}

\newcommand{\vx}{\mathbf{x}}
\newcommand{\vz}{\mathbf{z}}

\newcommand{\vb}{\boldsymbol{\beta}}

\newcommand{\vl}{\boldsymbol{\lambda}}

\definecolor{tablerowcolor}{rgb}{0.9, 0.95, 0.98} 
\definecolor{highlightcolor}{rgb}{0.9, 1.0, 1.0} 

\title{Latent Space Factorization in LoRA}

\author{%
    Shashi Kumar$^{1, 2}$ \quad Yacouba Kaloga$^{1}$ \quad John Mitros$^{1}$ \quad Petr Motlicek$^{1, 3}$ \quad Ina Kodrasi$^{1}$\\
    $^{1}$Idiap Research Institute, Switzerland \\
    $^{2}$EPFL, Switzerland \quad
    $^{3}$BUT, Czech Republic \\
    \texttt{\{shashi.kumar, yacouba.kaloga, petr.motlicek, ina.kodrasi\}@idiap.ch} \\
    \texttt{john.mitross@gmail.com}
}

\begin{document}

\maketitle

\begin{abstract}
Low-rank adaptation (LoRA) is a widely used method for parameter-efficient finetuning.
However, existing LoRA variants lack mechanisms to explicitly disambiguate task-relevant information within the learned low-rank subspace, potentially limiting downstream performance. 
We propose Factorized Variational Autoencoder LoRA (FVAE-LoRA), which leverages a VAE to learn two distinct latent spaces.
Our novel Evidence Lower Bound formulation explicitly promotes factorization between the latent spaces, dedicating one latent space to task-salient features and the other to residual information.
Extensive experiments on text, audio, and image tasks demonstrate that FVAE-LoRA consistently outperforms standard LoRA. 
Moreover, spurious correlation evaluations confirm that FVAE-LoRA better isolates task-relevant signals, leading to improved robustness under distribution shifts.
Our code is publicly available at: \url{https://github.com/idiap/FVAE-LoRA}
\end{abstract}

\section{Introduction}
\label{sec:introduction}

Foundation models have become ubiquitous across modalities such as vision~\cite{radford2021learning, kirillov2023segment, wu2020visual, rombach2022high}, audio~\cite{baevski2020wav2vec,whisper}, and text~\cite{brown2020language, grattafiori2024llama}. Recent state-of-the-art results are predominantly achieved by fine-tuning these large pre-trained models. Among various parameter-efficient fine-tuning (PEFT) strategies~\cite{houlsby2019parameter, liu2021gpt, lester2021power, hu2022lora}, \emph{Low-Rank Adaptation (LoRA)}~\cite{hu2022lora} has emerged as a particularly efficient approach. In LoRA, the original weight matrices \( \mathbf{W} \in \mathbb{R}^{k \times d} \) are kept frozen, and trainable low-rank matrices \( \mathbf{A} \in \mathbb{R}^{r \times d} \) and \( \mathbf{B} \in \mathbb{R}^{k \times r} \) are introduced, with \( r \ll min(d, k) \), such that the adapted weights become \( \mathbf{W} + \mathbf{B} \mathbf{A} \). This technique significantly reduces memory and computational requirements, while achieving performance comparable to full fine-tuning~\cite{hu2022lora,huang2025hira}.

Despite the remarkable performance shown by LoRA across a plethora of downstream tasks and modalities, we identify a potential limitation: the standard LoRA update mechanism lacks an explicit way to ensure that the learned low-rank subspace \( \text{Im}(\mathbf{A}) \) primarily captures task-salient information. The projection \( \mathbf{A} \vx \) (where \( \vx \) is the input activation) is learned implicitly through gradient descent on the task objective. While effective, this process does not inherently guarantee that \( \mathbf{A} \) isolates features crucial for the downstream task from potentially irrelevant or even detrimental information retained from pre-training. This lack of explicit control over the content of the low-rank update is pertinent.
While the hypothesis that fine-tuning primarily involves low-rank updates provides a strong theoretical underpinning for LoRA~\cite{aghajanyan2020intrinsic}, empirical evidence suggests nuances. Recent studies have shown that standard LoRA can still underperform full fine-tuning in certain scenarios~\cite{hu2022lora, liu2024dora}.
This suggests that simply constraining the update to be low-rank might not be sufficient; the task-relevant signal encoded within that low-rank adaptation is critical for achieving optimal downstream performance. Existing LoRA variants do not offer a principled mechanism to explicitly disentangle and prioritize task-relevant information within the learned update.

\begin{figure}[t] 
  \centering 
  \includegraphics[width=0.85\columnwidth]{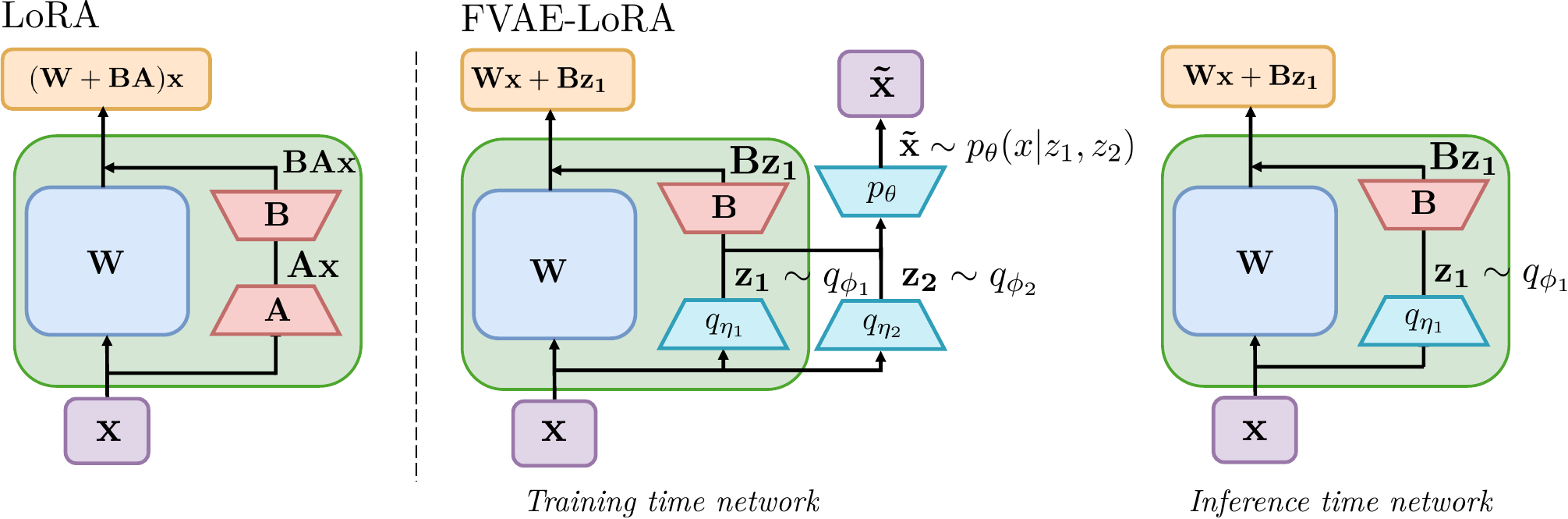} 
  \caption{Comparison between LoRA and the proposed FVAE-LoRA. During training, FVAE-LoRA factorizes the latent space into two components, $\vz_1$ and $\vz_2$, where only the task-salient latent factor $\vz_1$ is propagated downstream. At inference, only the encoder corresponding to $\vz_1$ is required.}
  \label{fig:fvae_lora} 
\end{figure}

To address this limitation and enable explicit control over the information captured within the low-rank update, we propose \textbf{Factorized Variational Autoencoder LoRA (FVAE-LoRA)}. Our approach integrates a Variational Autoencoder (VAE) framework directly into the LoRA parameterization. Crucially, unlike standard VAEs, FVAE learns \textbf{two distinct latent spaces}, denoted by \( \vz_1 \) and \( \vz_2 \) (see Figure~\ref{fig:fvae_lora}). We introduce a novel Evidence Lower Bound (ELBO) formulation specifically designed to promote \textbf{factorization} between these two spaces during training. This objective encourages the model to encode task-salient information, critical for downstream performance, primarily within the first latent space \( \vz_1 \), while relegating residual information necessary for accurate reconstruction (as required by the FVAE objective) to the second latent space \( \vz_2 \). During the forward pass for the downstream task, only samples drawn from the task-salient latent space \( \vz_1 \) are utilized to generate the effective low-rank adaptation matrix \( \mathbf{A} \). This mechanism allows FVAE-LoRA to explicitly select and leverage the most relevant learned features for the target task, while isolating potentially less useful or confounding variations within \( \vz_2 \).

Our main contributions can be summarized as follows:
\begin{itemize}
    \item \textbf{A Novel PEFT Method (FVAE-LoRA):} We propose FVAE-LoRA, integrating a VAE with factorized latent spaces (\(\vz_1\), \(\vz_2\)) into the LoRA framework to explicitly disentangle task-salient information (\(\vz_1\)) from residual information (\(\vz_2\)).
    \item \textbf{Factorizing ELBO Formulation:} We introduce a novel ELBO objective specifically designed to enforce this factorization between the two latent spaces during training.
    \item \textbf{Strong Empirical Performance:} We demonstrate through extensive experiments on diverse image, text, and audio benchmarks that FVAE-LoRA consistently outperforms LoRA. 
    \item \textbf{Empirical Validation of Robustness:} We empirically validate, using targeted spurious-correlation experiments, that the task-salient latent space \( \vz_1 \) indeed captures task-critical information, leading to a robust performance even on challenging examples designed to mislead standard LoRA.
\end{itemize}
\section{Related Work}
\label{sec:related_works}
We position FVAE-LoRA relative to PEFT methods, specifically LoRA variants, and techniques for latent space factorization in VAEs.

\noindent \textbf{PEFT} methods adapt large pre-trained models with minimal trainable parameters, overcoming the costs of full fine-tuning. Common approaches include inserting Adapter modules \cite{houlsby2019parameter}, optimizing continuous prompts or prefixes \cite{liu2021gpt, li2021prefixtuningoptimizingcontinuousprompts}, or tuning only bias terms \cite{zaken2022bitfitsimpleparameterefficientfinetuning}. LoRA \cite{hu2022lora} is a prominent PEFT technique that injects trainable low-rank matrices (\( \Delta \mathbf{W} = \mathbf{B} \mathbf{A}  \), rank \( r \ll min(d, k) \)) into the model layers. Its efficiency and performance have led to wide adoption \cite{aghajanyan2020intrinsic}. FVAE-LoRA builds upon the LoRA framework, aiming to enhance its effectiveness.

\noindent \textbf{LoRA Variations.} \quad
Several methods have extended LoRA. AdaLoRA \cite{zhang2023adalora} adaptively allocates rank budgets. DoRA \cite{liu2024dora} decouples weight magnitude and direction, applying LoRA to the latter. LoRA+ \cite{hayou2024loraefficientlowrank} adjusts LoRA's optimization by using different learning rates for its two low-rank matrices.
PiSSA \cite{meng2024pissa} focuses on initializing the LoRA matrices in a way that better approximates full fine-tuning updates, typically setting \( \mathbf{A} = \mathbf{0} \) and initializing \( \mathbf{B} \) based on principal components of the gradient.
RS-LoRA \cite{kalajdzievski2023rank} aims to stabilize training and prevent rank collapse or excessive growth by incorporating regularization related to the singular values of the update matrix.
Other contributions combine LoRA with quantization for further compression \cite{dettmers2023qlora,xu2023qa}.
These variants primarily modify the update's structure, optimization, or compression. Our work differs by focusing on the \emph{semantic content} of the update. FVAE-LoRA introduces a VAE with factorized latent spaces (\(\vz_1, \vz_2\)) and a novel ELBO to explicitly separate task-salient information (\(\vz_1\)) used for the update from residual information (\(\vz_2\)), thereby controlling the information encoded in the low-rank adaptation.

\noindent \textbf{Factorization and Disentanglement in VAEs.} \quad
VAEs \cite{kingma2022autoencodingvariationalbayes} learn latent representations by maximizing the ELBO. Significant research focuses on learning \emph{disentangled} representations, where latent dimensions capture independent factors of data variation \cite{locatello2018challenging}. Techniques often modify the ELBO, such as $\beta$-VAE \cite{higgins2017beta}, FHVAE~\cite{hsu2017unsupervised}, FactorVAE \cite{kim2019disentanglingfactorising}, TCVAE \cite{chen2019isolatingsourcesdisentanglementvariational}, and DIP-VAE \cite{kumar2018variationalinferencedisentangledlatent}, or employ annealing strategies \cite{burgess2018understandingdisentanglingbetavae}. While we use a VAE and aim for factorization, our goal is distinct. We do not seek to disentangle underlying data factors. Instead, FVAE-LoRA employs a novel ELBO to factorize the latent space specifically for PEFT: separating information crucial for the \emph{downstream task} (\(\vz_1\)) from other sources of variation needed for reconstruction (\(\vz_2\)). This task-conditional factorization within the LoRA update mechanism represents the core novelty of our VAE application.

\section{Method}
\label{sec:method}

In this section, we present our low-rank adaptation approach based on a VAE. We begin by briefly reviewing the standard VAE framework, and then introduce our proposed Factorized VAE (FVAE). We highlight key properties of this formulation and conclude by describing how it enables the construction of an efficient low-rank adaptation model.

\subsection{Variational Autoencoder Objective}

Consider a dataset \( X \in \mathbb{R}^{n \times d} \), where observations are generated according to the process \( \vx \sim p_{\theta}(\vx |\vz) \), with \( \vz \) a latent variable. The goal is to recover a latent representation \( \vz \) that explains the observations \( \vx \), which requires computing the posterior \( p_{\theta}(\vz |\vx) \). As this posterior is generally intractable, we introduce an approximate distribution \( q_{\phi}(\vz |\vx) \). It can be shown (see Appendix~\ref{appendix:elbo_derivation}) that the log-likelihood admits a lower bound, known as the \emph{Evidence Lower Bound (ELBO)}:
\begin{align}
\mathcal{L}_{\theta, \phi}^{\text{VAE}}(\vx) = \mathbb{E}_{\vz \sim q_\phi(\vz |\vx)} \left[ \log p_\theta(\vx |\vz) \right] - D_{\mathrm{KL}}\left(q_\phi(\vz |\vx) \,\|\, p(\vz)\right).
\label{eq:elbo}
\end{align}
The ELBO is used as a tractable surrogate objective for maximizing \( \log p_{\theta}(\vx) \). The first term encourages accurate reconstruction, while the second term regularizes the latent space by aligning the approximate posterior with the prior \( p(\vz) \).

\subsection{Factorized Variational Autoencoder Objective}

The primary goal of FVAE is to factorize the information contained in \( \vx \) such that it is represented by two independent latent variables \( \vz_1 \) and \( \vz_2 \). 
This factorization is learned jointly with a downstream task loss applied specifically to \( \vz_1 \), which guides the decomposition by encouraging \( \vz_1 \) to capture task-relevant information, while \( \vz_2 \) absorbs the remaining variability. Classical VAEs serve as a natural starting point to build such a model.



\subsubsection{Preliminaries}
The derivation of the classical VAE can be extended by assuming that \( \vx \) arises from a generative process involving two independent latent variables $\vz_1$ and $\vz_2$, with \( p(\vz_1, \vz_2) = p_1(\vz_1)\, p_2(\vz_2) \). Additionally, we assume that the approximate posterior factorizes as \( q_\phi(\vz_1, \vz_2 |\vx) = q_{\phi_1}(\vz_1 |\vx)\, q_{\phi_2}(\vz_2 |\vx) \). Considering $\mathbf{z}_{1} \sim q_{\phi_{1}}(\mathbf{z}_{1} \vert \mathbf{x})$ and $\mathbf{z}_{2} \sim q_{\phi_{2}}(\mathbf{z}_{2} \vert \mathbf{x})$, the ELBO is given by
\begin{align}
\label{eq:elbonaive}
\mathcal{L}_{\theta, \phi}^{\text{VAE2LAT}}(\vx) = \underset{\vz_1, \vz_2}{\mathbb{E}} \left[\log p_\theta(\vx |\vz_1, \vz_2)\right] 
- D_{\mathrm{KL}}\left(q_{\phi_1}(\vz_1 |\vx) \,\|\, p_1(\vz_1)\right) - D_{\mathrm{KL}}\left(q_{\phi_2}(\vz_2 |\vx) \,\|\, p_2(\vz_2)\right).
\end{align}


This objective mirrors the standard VAE but extends it to the multi-latent setting. However, even though both the prior and the variational posterior are factorized, the model is not explicitly encouraged to selectively assign information to \( \vz_1 \) or \( \vz_2 \).



\subsubsection{FVAE}

To  promote factorization, we introduce a regularization term that penalizes the similarity between \( q_{\phi_2}(\vz_2 | \vx) \) and the uninformative prior \( p_1(\vz_1) \). Since \( q_{\phi_1}(\vz_1 | \vx) \) is encouraged to align with \( p_1 \), this term prevents \( q_{\phi_2} \) from encoding information in the same region of the latent space.
Incorporating this into Equation~\eqref{eq:elbonaive}, we obtain the objective

\begin{equation}
\max_{\theta, \phi_1, \phi_2} \; 
\mathcal{L}_{\theta, \phi}^{\text{VAE2LAT}}(\vx) 
+  \mathbb{E}_{\vz_1, \vz_2} \left[ \log \frac{q_{\phi_2}(\vz_2 | \vx)}{p_1(\vz_1)} \right].
\label{eq:dkl_q_p_naive}
\end{equation}



To clarify the role of each component in Equation~\eqref{eq:dkl_q_p_naive} and relate them to familiar VAE structures, we reorganize the objective using straightforward algebraic manipulations. In doing so, we isolate the standard reconstruction and KL divergence terms, and separate out the new cross-prior regularizer. Introducing scalar constants $\alpha$, $\beta$ and $\delta$ allows us to balance the influence of these components, yielding the structured objective
\begin{align}
\label{eq:elbo-fvae}
\mathcal{L}_{\theta, \phi}^{\text{FVAE}}(\vx) &=  
\alpha \underset{\vz_1, \vz_2}{\mathbb{E}} \left[\log p_\theta(\vx | \vz_1, \vz_2)\right]   
- \beta D_{\mathrm{KL}}\left(q_{\phi_1}(\vz_1 | \vx) \,\|\, p_1(\vz_1)\right ) 
+ \delta \, \underbrace{\underset{\vz_2, \vz_1}{\mathbb{E}} \log \frac{p_2(\vz_2)}{p_1
(\vz_1)}}_{\Gamma}.
\end{align}

The second term correspond to the $D_{KL}$ in the \( \beta \)-VAE objective, ensuring that the main latent variable \( \vz_1 \) captures the relevant information for reconstructing \( \vx \) while remaining close to its prior. The third term, \( \Gamma \), acts as a repulsive regularizer, encouraging the second component \( \vz_2 \) to decouple from \( \vz_1 \). Note that, a priori, we could fix \( \alpha = 1 \) and use only \( \beta \) and \( \delta \) to weight the contributions of all the terms. However, we prefer to use all three, as it will make the interpretation of each contribution clearer later on.



\subsection{Mechanism of the $\Gamma$ modulator}
\label{sec:geom_interp_general}


$\Gamma$ introduces an indirect interaction between the two encoders by modulating their alignment with their respective priors. Rather than enforcing separation through a direct divergence between posteriors, it shifts their latent support via prior-based regularization. To analyze the effect of $\Gamma$, we first rewrite it as the sum of a mismatch term and a discrepancy term, i.e.,
\begin{equation}
    \Gamma  =   \underbrace{\mathbb{E}_{\vz_2 \sim q_{\phi_2}}\!\bigl[\log p_2 - \log p_1\bigr]}_{\text{mismatch: }\Lambda} + 
       \underbrace{\!\bigl[\mathbb{E}_{\vz_2 \sim q_{\phi_2}}\log p_1 - \mathbb{E}_{\vz_1 \sim q_{\phi_1}}\log p_1\bigr]}_{\text{discrepancy: }\Delta},
\end{equation}
where the mismatch term can be further equivalently expressed as a difference of KLs, i.e., $\Lambda = D_{KL} (q_{\phi_2} || p_1) - D_{KL} (q_{\phi_2} || p_2)$.
This decomposition reveals a meaningful structure, as outlined in the following.

Maximizing the mismatch encourages \( q_{\phi_2} \) to align with its prior \( p_2 \). This mirrors the behavior expected in a two-variable standard VAE, where each encoder is regularized toward its respective prior. As a result, we retain effective control over the behavior of \( q_{\phi_2} \), providing a structural safeguard against degenerate or unconstrained posterior collapse. In contrast, it disincentivizes \( q_{\phi_2} \) from aligning with the prior \( p_1 \). Consequently, \( q_{\phi_2} \) is encouraged to preserve or discover features and structures that are distinct from, and not merely reflections of, the assumptions embedded within \( p_1 \). The mismatch term also highlights that the two priors \( p_1 \) and \( p_2 \) should be different, but still partially overlapping. If they are identical, some terms will simply cancel out, and if they are too far apart, the separation becomes trivial, resulting in no fruitful competition for occupying the latent space. Since the priors are usually Gaussian with variance 1, this competition is parameterized by \( |\mu_1 - \mu_2| \); the effect of the mismatch is null when this parameter is null.


In addition, we can demonstrate (see Appendix~\ref{app:delta_bound}) that the discrepancy \( \Delta \) is bounded by a term depending on the 2-Wasserstein distance, provided that the Hessian \( \|\nabla^2 \log p_1\| \le L \) is bounded. In practice, \( p_1 \) is typically a standard normal \( \mathcal{N}(0, I) \), and \( q_{\phi_1} \) is a diagonal Gaussian. Under these assumptions, the bound becomes:
\[
\Delta \leq \frac{L}{2} \mathcal{W}_2^2(q_{\phi_1}, q_{\phi_2}) + \sqrt{\sum_j \mu_j^2 + \sigma_j^2} \cdot \mathcal{W}_2(q_{\phi_1}, q_{\phi_2}),
\]
where \( \mu_j \) and \( \sigma_j^2 \) are the parameters of \( q_{\phi_1} \). Since \( q_{\phi_1} \) is typically optimized to approximate \( p_1 \), the square-root term remains bounded in most settings. Both terms in the bound grow with \( \mathcal{W}_2(q_{\phi_1}, q_{\phi_2}) \), making \( \Delta \) an effective surrogate for inducing Wasserstein repulsion. In particular, maximizing \( \Delta \) increases \( \mathcal{W}_2(q_{\phi_1}, q_{\phi_2}) \), driving the two encoders apart in a geometrically meaningful way.

\subsection{FVAE-LoRA}
\label{sec:fvae_lora_impl}

Building upon the FVAE framework, we leverage its ability to split the latent space to gain finer control over the representation, ultimately achieving better performance. To accomplish this, we proceed as illustrated in Figure~\ref{fig:fvae_lora}.

For each targeted linear layer, we train an FVAE simultaneously with the downstream task, aiming to replace the $\mathbf{A}$ matrices used in classical LoRA (see the left side of the figure). During training, the input to the target layer is fed into the FVAE to compute a reconstruction loss based on that input. In parallel, the latent embedding $\mathbf{z}_1$ produced by the encoder $q_{\phi_1}$ is passed through a learned matrix $\mathbf{B}$ and added to the output of the frozen base weights $\mathbf{W}$. This yields the output $\mathbf{W}\mathbf{x} + \mathbf{B} \mathbf{z}_1$. At inference time, only $q_{\phi_1}$ is used to produce the output, either by sampling from it or by taking the mean of the distribution. Note that while we propose using FVAE with LoRA, the method is generic in the sense that it can be applied to give latent space control to any explicit LoRA method. In summary, the loss to be optimized in the proposed FVAE-LoRA approach is given by
\begin{equation}
\label{eq:to_refer_downstream}
   \min_{\phi, \theta} \mathcal{L}_{\text{downstream-task}} -  \vl \sum_{l \in \text{layer}} \mathcal{L}_{\theta, \phi}^{\text{FVAE}}(\vx_l),
\end{equation}
with $\vl$ being the hyper-parameter vector of weights assigned to the FVAE loss in each layer.

\textbf{In practice:} Both $q_{\phi_1}$ and $q_{\phi_2}$ are parameterized as diagonal Gaussian distributions, with their means and variances learned by neural networks. The reconstruction term $p_{\theta}(\mathbf{x} | \mathbf{z}_1, \mathbf{z}_2)$ is also parameterized by a neural network. The prior $p_1 =  \mathcal{N}(\mathbf{0},  \mathbf{I})$ is a standard normal distribution, while $p_2$ is empirically chosen to be centered at $\mathbf{1.5}$.
\CR{
The intuition is to give the two priors distinct non-overlapping "location" in the latent space to initialize and encourage separation. By setting $\mu_1$ at $0$ and $\mu_2$ at $1.5$, we provide a clear signal for the repulsive regularizer to push the posteriors apart. See additional insights in Appendices \ref{appendix:early_attempts} and \ref{appendix:additional_insights}.
}

\section{Experimental Results}
\label{sec:experimental_results}
%
\noindent \textbf{Motivation.} \quad The objective of the experimental evaluation is two fold. First, we aim to comprehensively evaluate FVAE-LoRA by comparing its performance against standard LoRA and its relevant variants across diverse image, text, and audio tasks. The specific selection of relevant variants for each domain is detailed within the respective modality subsections, guided by the aim to provide the most insightful and relevant benchmarks for each specific context. Second, we seek to empirically validate that FVAE-LoRA learns more robust representations by preferentially encoding task-salient information in $\vz_1$.

\noindent \textbf{Overall Setup.} \quad
To ensure fair comparisons of parameter efficiency for the core adaptation mechanism, the LoRA rank \( r \) is set to \(16\) for all LoRA-based methods throughout our experiments. This rank also corresponds to the dimensionality of the task-salient latent space \( \vz_1 \) in FVAE-LoRA. All LoRA-based baselines as well as FVAE-LoRA are applied to the query and key matrices within the transformer models. Detailed hyperparameter settings for FVAE-LoRA, including the balancing coefficients $\alpha$, $\beta$ and $\delta$, learning rates, and specific VAE architectural choices for each task, are provided in Appendix~\ref{appendix:hparams_details}. \CR{We also provide a practical guide for selecting the key factorization hyperparameters, $\beta$ and $\delta$, in Appendix \ref{appendix:hparam_guide}.}

\subsection{Efficacy of FVAE-LoRA for Various Downstream Tasks}
\label{sec:exp_modalities}

\subsubsection{Image Tasks}
\label{sec:exp_image}
\noindent \textbf{Datasets.} \quad
We evaluate FVAE-LoRA on six diverse image classification datasets: DTD~\cite{cimpoi2014describing}, EuroSAT~\cite{helber2019eurosat}, GTSRB~\cite{houben2013detection}, RESISC45~\cite{cheng2017remote}, SUN397~\cite{xiao2010sun}, and SVHN~\cite{netzer2011reading}. These datasets span various image types, domains, and complexities.

\noindent \textbf{Implementation Details.} \quad
The pre-trained Vision Transformer (ViT-B/16)~\cite{wu2020visual} serves as the backbone model for all image classification tasks. We compare FVAE-LoRA against full fine-tuning (Full FT) and several LoRA variants, i.e., standard LoRA~\cite{hu2022lora}, PiSSA~\cite{meng2024pissa}, rsLoRA~\cite{kalajdzievski2023rank}, DoRA~\cite{liu2024dora}, and OLoRA~\cite{buyukakyuz2024olora}. This broad selection of LoRA variants represents established PEFT methods for image classification.
The evaluation metric is top-1 accuracy.
Detailed hyperparameters can be found in \ref{appendix:hparams_details_image}.

\begin{table}[h!]
\centering
\caption{Fine-tuning results of ViT-B/16 on image classification tasks. We fine-tune ViT-B/16 using full fine-tuning and LoRA variants across DTD, EuroSAT, GTSRB, RESISC45, SUN397, and SVHN datasets. \textbf{Bold} indicates the highest performance, while \underline{underline} represents the second-highest performance.}
\label{tab:finetuning_results_image}
\resizebox{\columnwidth}{!}{
\begin{tabular}{@{}lc|cccccc|c@{}}
\toprule
Method    & \textbf{Params (\%)} & \textbf{DTD} & \textbf{EuroSAT} & \textbf{GTSRB} & \textbf{RESISC45} & \textbf{SUN397} & \textbf{SVHN} & \textbf{Average} \\
\midrule
Full FT   & - & \underline{78.12$\pm$0.59} & \textbf{98.30$\pm$0.47} & \textbf{98.85$\pm$0.14} & \textbf{94.35$\pm$0.54} & 69.34$\pm$0.59 & \textbf{97.34$\pm$0.03} & \underline{89.38} \\
\midrule
LoRA      & 0.7240 & 74.65$\pm$1.08 & 97.28$\pm$0.36 & 96.95$\pm$0.56 & 90.11$\pm$0.53 & 71.11$\pm$0.07 & 94.22$\pm$0.14 & 87.39 \\
PiSSA     & 0.7240 & 74.22$\pm$1.69 & 97.33$\pm$0.31 & 96.95$\pm$1.28 & 89.82$\pm$0.37 & 69.09$\pm$0.19 & 94.83$\pm$0.73 & 87.04 \\
rsLoRA    & 0.7240 & 72.23$\pm$1.00 & 97.48$\pm$0.21 & 96.63$\pm$0.83 & 88.04$\pm$0.30 & 67.69$\pm$0.32 & 93.81$\pm$0.77 & 85.98 \\
DoRA      & 0.7451 & 75.74$\pm$1.91 & 97.28$\pm$0.80 & 97.27$\pm$0.44 & 91.72$\pm$1.17 & \underline{71.53$\pm$0.25} & 96.41$\pm$0.72 & 88.32 \\
OLoRA    & 0.7240 & 72.23$\pm$0.40 & 96.62$\pm$1.06 & 97.08$\pm$0.46 & 88.94$\pm$0.45 & 69.64$\pm$0.43 & 94.86$\pm$0.30 & 86.63 \\
\midrule
\rowcolor{cyan!20} 
FVAE-LoRA  & 0.7311 & \textbf{78.19$\pm$0.68} & \underline{97.78$\pm$0.15} & \underline{97.98$\pm$0.56} & \underline{93.57$\pm$0.22} & \textbf{73.14$\pm$0.21} & \underline{96.55$\pm$0.05} & \textbf{89.53} \\
\bottomrule
\end{tabular}
}
\end{table}
\noindent \textbf{Results.} \quad
The effectiveness of FVAE-LoRA for image classification is shown in Table~\ref{tab:finetuning_results_image}. FVAE-LoRA achieves an average accuracy of 89.53\% across six diverse datasets, outperforming LoRA and surpassing variants such as DoRA, all within a comparable inference-time parameter budget.

Notably, FVAE-LoRA's average performance slightly surpasses that of full fine-tuning (89.38\%). This result suggests that the structured latent factorization inherent to FVAE-LoRA can guide the model towards learning highly effective adaptations. By explicitly encouraging the disentanglement of task-salient information within $\vz_1$, FVAE-LoRA might be more adept at focusing the ViT backbone on critical visual features for the downstream task, potentially mitigating the risk of overfitting to spurious correlations or less generalizable patterns that can sometimes affect full fine-tuning on these datasets.
On challenging datasets such as DTD (characterized by fine-grained textures) and SUN397 (complex scenes), FVAE-LoRA particularly excels, achieving the highest scores and outperforming full fine-tuning. For instance, on SUN397, FVAE-LoRA demonstrates a clear advantage, indicative of its capacity to distill critical visual cues for complex recognition tasks. While full fine-tuning outperforms all LoRA variants on datasets like EuroSAT and GTSRB, FVAE-LoRA consistently stands as the leading or a highly competitive PEFT method, often closing the gap significantly (e.g., achieving 97.78\% on EuroSAT, closely trailing Full FT's 98.30\%).

The presented results show that FVAE-LoRA is able to learn highly effective low-rank updates through a principled approach to information selection. 

\subsubsection{Text Tasks}
\label{sec:exp_text}
\noindent \textbf{Datasets.} \quad
For natural language tasks, we use two benchmark categories:
\begin{enumerate}
    \item \textbf{Commonsense Reasoning}: Training is done on a predefined corpus~\cite{hu2023llm}\footnote{\url{https://github.com/AGI-Edgerunners/LLM-Adapters/tree/main/dataset}} of query-answer pairs, and the evaluation set includes seven sub-tasks: PIQA~\cite{piqa} (physical commonsense), SIQA~\cite{siqa} (social interaction understanding), ARC-c and ARC-e~\cite{arc} (science question answering), OBQA~\cite{mihaylov2018can} (multi-hop reasoning over facts), HellaSwag~\cite{zellers2019hellaswag} (commonsense natural language inference)), and WinoGrande~\cite{winogrande} (fill-in-the-blank).
    \item \textbf{GLUE Benchmark}: A subset of the GLUE~\cite{wang2019glue} is used, comprising SST2 (sentiment analysis), CoLA (linguistic acceptability), QNLI (question-answering NLI), MRPC (paraphrase detection), RTE (textual entailment), STSB (semantic textual similarity), and WNLI (coreference resolution).
\end{enumerate}

\noindent \textbf{Implementation Details.} \quad
We employ Llama-3-8B~\cite{grattafiori2024llama} for the commonsense reasoning tasks and roberta-base~\cite{liu2019roberta} for the GLUE benchmark tasks.
For commonsense reasoning tasks, we compare against Prompt Tuning~\cite{lester2021power}, P-Tuning~\cite{liu2021gpt, liu-etal-2022-p}, standard LoRA, and HiRA~\cite{huang2025hira}. For completeness, we also present the performance of ChatGPT taken from \cite{liu2024dora}.
Considering the computational cost of LLM fine-tuning, our LoRA-based comparisons focus on standard LoRA and HiRA, as HiRA has recently demonstrated strong performance, offering a relevant and challenging benchmark in this setting.
For roberta-base on GLUE, comparisons are made against Full FT and standard LoRA. This allows for a direct assessment of FVAE-LoRA's parameter efficiency relative to the crucial full fine-tuning upper bound and the widely adopted LoRA baseline.
Evaluation uses accuracy for commonsense tasks following~\cite{huang2025hira} and standard GLUE metrics (Matthews Correlation for CoLA, Pearson Correlation for STSB, Accuracy for the rest).
Detailed hyperparameters can be found in \ref{appendix:hparams_details_text}.

\begin{table}[h!]
\centering
\caption{Accuracy comparison among various PEFT methods on commonsense reasoning datasets for Llama-3-8B. \textbf{Bold} indicates the best performance, while \underline{underline} represents the second-best performance. ChatGPT performance values are taken from \cite{liu2024dora}, whereas Prompt Tuning and P-Tuning from \cite{huang2025hira}.}
\label{tab:llama3_commonsense_modified}
\resizebox{\columnwidth}{!}{
\begin{tabular}{@{}l l c|c c c c c c c|c@{}}
\toprule
\textbf{Model} & \textbf{Method} & \textbf{Params (\%)} & \textbf{PIQA} & \textbf{SIQA} & \textbf{ARC-c} & \textbf{ARC-e} & \textbf{OBQA} & \textbf{HellaSwag} & \textbf{WinoGrande} & \textbf{Average} \\
\midrule
ChatGPT & - & - & 85.40 & 68.50 & 79.90 & 89.80 & 74.80 & 78.50 & 66.10 & 77.57 \\
\midrule
\multirow{5}{*}{Llama-3-8B} & Prompt Tuning & 0.0010 & 45.05 & 36.13 & 31.57 & 32.74 & 29.20 & 14.01 & 50.12 & 34.12 \\
 & P-Tuning & 0.6240 & 11.64 & 8.19 & 7.42 & 8.63 & 9.60 & 1.77 & 37.65 & 12.13 \\
 & LoRA  & 0.0848 & 80.74 & 75.59 & 67.58 & 82.11 & 75.20 & 85.73 & 77.82 & 77.82 \\
 & HiRA  & 0.0848 & \underline{88.63} & \underline{80.40} & \textbf{81.66} & \textbf{93.56} & \textbf{87.20} & \underline{94.48} & \underline{85.87} & \underline{87.40} \\
 \rowcolor{cyan!20} 
     & FVAE-LoRA & 0.0850 & \textbf{88.96} & \textbf{81.58} & \underline{81.06} & \underline{92.72} & \underline{86.20} & \textbf{95.30} & \textbf{88.95} & \textbf{87.82} \\ 
\bottomrule
\end{tabular}
}
\end{table}
\noindent \textbf{Results on Commonsense Reasoning using Llama-3-8B model.} \quad
Table~\ref{tab:llama3_commonsense_modified} reports the performance of FVAE-LoRA and baselines across seven commonsense reasoning benchmarks using the LLaMA-3-8B model. Our approach achieves the highest average accuracy of 87.82\%, outperforming both the strong HiRA baseline (87.40\%) and LoRA (77.82\%) under comparable inference-time parameter budgets.
These results indicate that FVAE-LoRA's strategy of factorizing latent information is particularly beneficial for complex reasoning tasks in LLMs. By explicitly guiding the $\vz_1$ latent space to capture task-salient semantic and contextual cues necessary for reasoning, FVAE-LoRA enables Llama-3-8B to make more accurate inferences. 

Notably, FVAE-LoRA demonstrates strong individual performances on tasks like HellaSwag (95.30\%) and WinoGrande (88.95\%), which require nuanced understanding of everyday situations and disambiguation. This suggests that the information isolated in $\vz_1$ is indeed critical for these types of reasoning, allowing the LLM to leverage its capabilities more effectively than with less structured adaptation techniques. The ability to improve upon already powerful models like Llama-3-8B with such parameter efficiency highlights the potential of FVAE-LoRA for targeted capability enhancement in large language models.

\begin{table}[h!]
    \centering
    \caption{Results of fine-tuning roberta-base using full fine-tuning and LoRA on a subset of the GLUE datasets. \textbf{Bold} indicates the best results, while \underline{underline} represents the second-best results.}
    \label{tab:glue_results_main}
    \resizebox{\textwidth}{!}{
    \begin{tabular}{l c | c c c c c c c | c}
        \toprule
        Method & \textbf{Params (\%)} &  \textbf{SST2} & \textbf{CoLA} & \textbf{QNLI} & \textbf{MRPC} & \textbf{RTE} & \textbf{STSB} & \textbf{WNLI} & \textbf{Average} \\
        \midrule
        Full FT & - & \textbf{94.77$\pm$0.25} & \textbf{62.43$\pm$1.16} & \textbf{91.97$\pm$0.05} & \underline{89.40$\pm$0.70} & \underline{79.53$\pm$1.31} & \textbf{90.30$\pm$0.29} & 56.30$\pm$0.00 & \underline{80.67} \\
        LoRA    & 0.4710 & 93.97$\pm$0.47 & 59.60$\pm$1.02 & \underline{91.87$\pm$0.19} & 88.73$\pm$0.61 & 77.87$\pm$0.74 & \underline{88.90$\pm$0.45} & \underline{57.73$\pm$2.03} & 79.81 \\
        \midrule
         \rowcolor{cyan!20} 
        FVAE-LoRA & 0.4759 & \underline{94.10$\pm$0.43} & \underline{60.37$\pm$0.49} & 91.63$\pm$0.34 & \textbf{89.53$\pm$0.97} & \textbf{79.90$\pm$0.85} & 88.60$\pm$0.16 & \textbf{64.33$\pm$2.38} & \textbf{81.21} \\
        \bottomrule
    \end{tabular}
    }
\end{table}
\noindent \textbf{Results on GLUE benchmark.} \quad
Table~\ref{tab:glue_results_main} presents the performance of FVAE-LoRA when adapting the roberta-base model on a subset of the GLUE benchmark. Our method achieves the highest average score (81.21), outperforming both full fine-tuning (80.67) and standard LoRA (79.81). Notably, FVAE-LoRA shows particular strength on tasks like MRPC and WNLI. This strong performance on roberta-base demonstrates that the benefits of FVAE-LoRA's explicit latent factorization are not confined to large-scale models like Llama-3-8B (as seen in commonsense reasoning tasks), but also translate effectively to smaller, yet widely utilized encoder models. The ability to enhance these more moderately-sized architectures suggests that FVAE-LoRA's principled approach to focusing adaptations via $\vz_1$ on task-critical linguistic features is robust across different model scales. 

\subsubsection{Audio Tasks}
\label{sec:exp_audio}
\textbf{Datasets.} \quad
We conduct automatic speech recognition (ASR) on the TIMIT acoustic-phonetic corpus~\cite{garofolo1993timit} for phoneme recognition. 

\begin{wraptable}[12]{r}{0.45\columnwidth} 
    \centering
    \caption{Fine-tuning results of Wav2Vec2-Large on the TIMIT speech recognition task using CTC loss. \textbf{Bold} indicates the best PER ($\downarrow$), \underline{underline} the second-best.}
    \label{tab:results_audio_wrapped}
    \vspace{-0.2cm} 
    \begin{tabular}{lc|c}
    \toprule
    Method    & \textbf{Params (\%)} & \textbf{TIMIT} \\
              & & PER ($\downarrow$) \\
    \midrule
    Full FT   & - & \textbf{7.48} \\
    LoRA      & 0.4961 &  9.38 \\
    \midrule
    \rowcolor{cyan!20}
    FVAE-LoRA  & 0.4999 & \underline{8.09} \\
    \bottomrule
    \end{tabular}
    \vspace{-0.2cm} 
\end{wraptable}

\noindent \textbf{Implementation Details.} \quad
The pre-trained Wav2Vec2-Large model~\cite{baevski2020wav2vec} serves as the backbone. Fine-tuning utilizes the Connectionist Temporal Classification loss~\cite{graves2006connectionist}. We compare against Full FT and standard LoRA. Performance is measured by Phoneme Error Rate (PER). Detailed hyperparameters are provided in Appendix~\ref{appendix:hparams_details_audio}.


\noindent \textbf{Results.} \quad
As shown in Table~\ref{tab:results_audio_wrapped}, FVAE-LoRA achieves a PER of 8.09 on TIMIT, outperforming standard LoRA and approaching the performance of full fine-tuning (7.48), demonstrating its effectiveness for ASR.

\subsection{Probing Latent Factorization via Spurious Correlation}
\label{sec:exp_spurious}
To empirically validate our hypothesis that FVAE-LoRA learns more robust representations by preferentially encoding task-salient information in $\vz_1$, we conduct experiments using datasets with controlled spurious correlations.
Spurious correlations occur when input features are statistically associated with target labels without a true causal link~\cite{qiu2024complexity, sreekumar2023spurious, pmlr-v119-sagawa20a}, potentially misleading models and hindering generalization, especially on out-of-distribution or minority-group data. Our aim is to assess whether FVAE-LoRA's disentanglement mechanism renders it more robust to such misleading cues compared to standard LoRA.

\begin{figure}[htbp]
    \centering
    \begin{subfigure}[b]{0.25\columnwidth}
        \centering
        \caption*{$\mathbf{x}$}
        \includegraphics[width=\columnwidth]{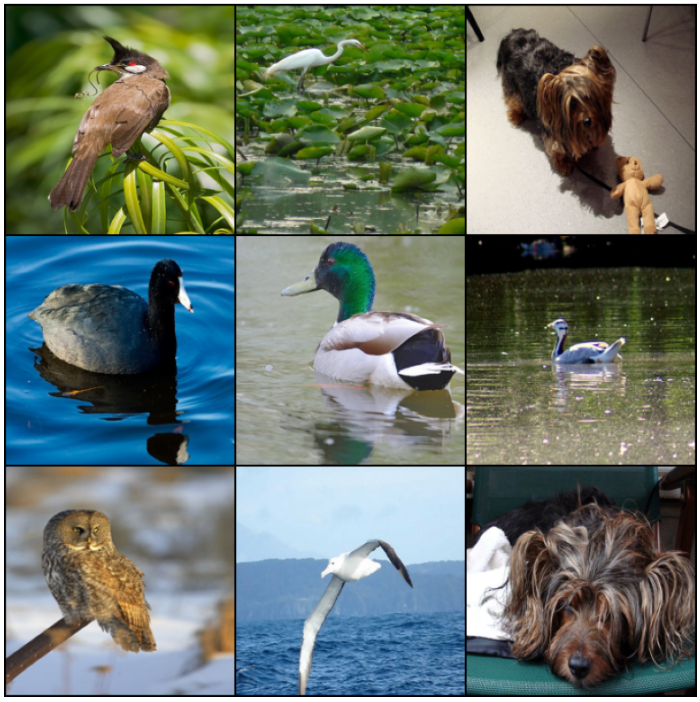} 
    \end{subfigure}
    \begin{subfigure}[b]{0.25\columnwidth}
        \centering
        \caption*{$\mathbf{x}_\text{core}$}
        \includegraphics[width=\columnwidth]{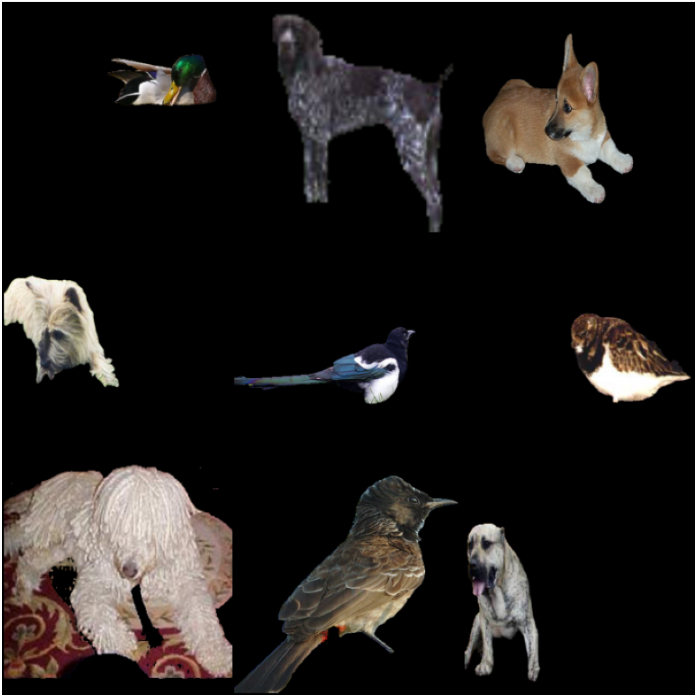} 
    \end{subfigure}
    \begin{subfigure}[b]{0.25\columnwidth}
        \centering
        \caption*{$\mathbf{x}_\text{spurious}$}
        \includegraphics[width=\columnwidth]{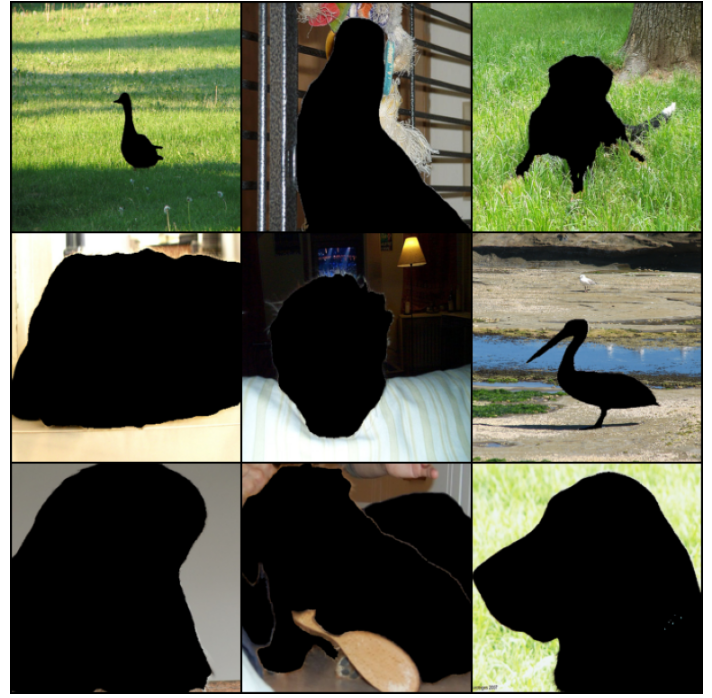} 
    \end{subfigure}
    \caption{Random samples drawn from the train split of the Animals dataset, illustrating an original image ($\mathbf{x}$), its core object features ($\mathbf{x}_\text{core}$), and its spurious background features ($\mathbf{x}_\text{spurious}$).}
\label{fig:spuco_animals}
\end{figure}

\noindent \textbf{Experimental Design.} \quad We leverage datasets where spurious attributes (e.g., background scene) are intentionally correlated with the true class labels (e.g., object category) during training. For example, a ``landbird" might predominantly appear against a "land" background, and a "waterbird" against "water". Effective factorization should enable the model to learn the true object category via \(\vz_1\), irrespective of the potentially misleading background. Figure~\ref{fig:spuco_animals} illustrates this concept, distinguishing between an input image ($\mathbf{x}$), its core features ($\mathbf{x}_{\text{core}}$), and its spurious features ($\mathbf{x}_{\text{spurious}}$).

\noindent \textbf{Datasets.} \quad Following prior works \citep{cui2024ameliorate, qiu2024complexity, sreekumar2023spurious, pmlr-v119-sagawa20a, pmlr-v119-srivastava20a, koh2021wilds}, we consider three standard benchmarks to introduce spurious correlations:
\textit{Waterbirds}~\cite{koh2021wilds}, where bird type (landbird vs. waterbird) is correlated with background (land vs. water);
\textit{CelebA}~\cite{koh2021wilds}, where a target attribute (e.g., blonde hair) might be correlated with another attribute (e.g., being female); and
\textit{Animals}~\cite{joshi2025challenges}, a larger-scale dataset derived from ImageNet~\cite{deng2009imagenet} with four animal classes spuriously correlated with background types (e.g., waterbirds with water, small dogs with indoor scenes).
These datasets are structured into groups based on combinations of true labels and spurious attributes, with varying majority-to-minority group ratios between training and test splits (details in Appendix~\ref{appendix:dataset_details_spurious} and Table~\ref{tab:spuco_datasets}).

\noindent \textbf{Implementation Details and Evaluation Metrics.} \quad
We adapt the ViT-B/16~\cite{wu2020visual} backbone using LoRA and our proposed FVAE-LoRA. Following common practice in literature~\cite{sagawa2020distributionally, creager2021environment, liu2021just}, performance is evaluated using three key metrics:
\begin{itemize}
    \item \textbf{Worst-Group Accuracy (WG):} Accuracy on the test subgroup where the model performs poorest, indicating robustness to spurious correlations and performance on minority groups.
    \item \textbf{Average Accuracy (AVG):} Standard overall accuracy on the test set.
    \item \textbf{Accuracy Disparity:} The absolute difference $\vert \text{WG} - \text{AVG} \vert$, quantifying the performance variation across groups. A smaller disparity suggests more uniform and equitable performance.
\end{itemize}

\begin{table}[h!]
    \centering
    \caption{Fine-tuning results of ViT-B/16 on spurious correlation benchmarks. We compare LoRA with FVAE-LoRA on ANIMALS (8 groups, 4 classes), WATERBIRDS (4 groups, 2 classes), and CELEBA (4 groups, 2 classes) datasets.}
    \label{tab:spurious_results_formatted_v2}
    \resizebox{\textwidth}{!}{%
    \begin{tabular}{l c | c c c c c c | c}
        \toprule
        \textbf{Method} & \textbf{Params (\%)} &
        \multicolumn{2}{c}{\textbf{ANIMALS}} &
        \multicolumn{2}{c}{\textbf{WATERBIRDS}} &
        \multicolumn{2}{c}{\textbf{CELEBA}} &
        \textbf{Disparity} \\
        \cmidrule(lr){3-4} \cmidrule(lr){5-6} \cmidrule(lr){7-8}
        & & 
        \textbf{WG} & \textbf{AVG} & 
        \textbf{WG} & \textbf{AVG} & 
        \textbf{WG} & \textbf{AVG} & 
        \textbf{$\vert$ WG - AVG $\vert$} \\ 
        \midrule
        LoRA & 0.7240 &
        54.79$\pm$8.08 & 88.20$\pm$1.17 & 
        75.49$\pm$0.9 & 90.39$\pm$0.78 & 
        40.00$\pm$3.54 & \textbf{96.09$\pm$0.02} & 
        34.8 \\ 
        \midrule
        \rowcolor{cyan!20} 
        FVAE-LoRA & 0.7311 &
        \textbf{62.0$\pm$4.83} & \textbf{89.55$\pm$0.96} & 
        \textbf{75.85$\pm$3.72} & \textbf{90.99$\pm$0.51} & 
        \textbf{43.33$\pm$6.68} & 95.77$\pm$0.18 & 
        31.71 \\ 
        \bottomrule
    \end{tabular}%
    }
\end{table}
\noindent \textbf{Results.} \quad
Table~\ref{tab:spurious_results_formatted_v2} summarizes the performance of standard LoRA, and FVAE-LoRA on the spurious correlation benchmarks.
Across all datasets, FVAE-LoRA consistently achieves higher WG and lower Accuracy Disparity compared to LoRA, while maintaining competitive AVG.
These findings strongly suggest that FVAE-LoRA is less susceptible to being misled by spurious features present in the training data. We attribute this enhanced robustness to the explicit factorization encouraged by our novel ELBO. By compelling \(\vz_1\) to capture genuinely task-relevant, causal features and relegating other variations, FVAE-LoRA learns a more robust adaptation. This leads to improved generalization, particularly on minority groups where spurious cues are often unreliable or reversed, thereby validating the intended robust learning mechanism of our proposed method.



\begin{table}[h!]
\centering
\caption{Ablation study comparing our FVAE-LoRA when fine-tuning ViT-B/16 to a two-latent-variable VAE (VAE2LAT, as defined in Eq.~\eqref{eq:elbonaive}), and $\beta$-VAE2LAT, the $\beta$-VAE version of VAELAT (where all the DKL terms are multiplied by 10). Results are presented on DTD, EuroSAT, GTSRB, RESISC45, SUN397, and SVHN. \textbf{Bold} indicates the highest results, while \underline{underlined} indicates the second-highest.}

\label{tab:finetuning_results_image_ablation_kl}
\resizebox{\columnwidth}{!}{%
\begin{tabular}{l|cccccc|c}
\toprule
Method    & \textbf{DTD} & \textbf{EuroSAT} & \textbf{GTSRB} & \textbf{RESISC45} & \textbf{SUN397} & \textbf{SVHN} & \textbf{Average} \\
\midrule
\rowcolor{cyan!20} 
FVAE-LoRA (Proposed) & \textbf{78.19$\pm$0.68} & \textbf{97.78$\pm$0.15} & \textbf{97.98$\pm$0.56} & \textbf{93.57$\pm$0.22} & \textbf{73.14$\pm$0.21} & \textbf{96.55$\pm$0.05} & \textbf{89.53} \\
\midrule
$\vb$-VAE2LAT ~ \eqref{eq:betavae2lat}  (where $\vb=10$)    & \underline{77.16$\pm$0.43} & \underline{96.86$\pm$0.15} & \underline{95.75$\pm$0.46} & \underline{89.46$\pm$0.13} & \underline{72.91$\pm$0.32} & \underline{91.58$\pm$0.69} & \underline{87.29} \\
\midrule
VAE2LAT ~ \eqref{eq:dkl_q_p_naive} & 75.96$\pm$0.80 & 96.64$\pm$0.59 & 94.38$\pm$0.94 & 88.42$\pm$0.32 & 71.68$\pm$0.54 & 91.50$\pm$0.58 & 86.43 \\
\bottomrule
\end{tabular}%
}
\end{table}
\subsection{Ablation Studies}
\label{sec:res_ablations}
To demonstrate the relevance of introducing the regularization term in Equation~(\eqref{eq:dkl_q_p_naive}), we replicate our image results using the two-variable VAE model~(\ref{eq:elbonaive}) and its equivalent for $\beta$-VAE with two latent variables (where the two KL divergences have been multiplied by 10; see Equation~\eqref{appendix:elbo3_derivation}). The results can be seen in Table~\ref{tab:finetuning_results_image_ablation_kl}. The baseline model performs the worst across all datasets. The $\beta$-VAE with two latent variables shows some improvement, but it is still outperformed by our proposed method.

\section{Conclusions}
We introduced Factorized Variational Autoencoder LoRA (FVAE-LoRA), a novel PEFT method designed to explicitly disentangle task-salient information within the LoRA framework. By employing a VAE with two latent spaces, $\vz_1$ (task-salient) and $\vz_2$ (residual), and a specialized ELBO, FVAE-LoRA ensures that the adaptive updates are primarily driven by task-critical features learned in $\vz_1$.
Our comprehensive evaluations on diverse text, audio, and image benchmarks demonstrated that FVAE-LoRA consistently surpasses standard LoRA in performance. Crucially, experiments on datasets with spurious correlations empirically confirmed that FVAE-LoRA's factorization leads to more robust representations, as evidenced by improved worst-group accuracy.
FVAE-LoRA highlights the potential of latent space factorization for enhancing parameter-efficient fine-tuning.

\section{Acknowledgments}
Shashi Kumar was partially supported by the EU Horizon 2020 project ELOQUENCE (grant number 101070558).
Yacouba Kaloga was partially supported by Swiss National Science
Foundation project no CRSII5\_202228 on Characterisation of motor speech disorders and processes.

\bibliographystyle{unsrt}
\bibliography{neurips_2025}

\begin{thebibliography}{10}

\bibitem{radford2021learning}
Alec Radford, Jong~Wook Kim, Chris Hallacy, Aditya Ramesh, Gabriel Goh, Sandhini Agarwal, Girish Sastry, Amanda Askell, Pamela Mishkin, Jack Clark, et~al.
\newblock Learning transferable visual models from natural language supervision.
\newblock In {\em ICML}, 2021.

\bibitem{kirillov2023segment}
Alexander Kirillov, Eric Mintun, Nikhila Ravi, Hanzi Mao, Chloe Rolland, Laura Gustafson, Tete Xiao, Spencer Whitehead, Alexander~C Berg, Wan-Yen Lo, et~al.
\newblock Segment anything.
\newblock In {\em ICCV}, 2023.

\bibitem{wu2020visual}
Bichen Wu, Chenfeng Xu, Xiaoliang Dai, Alvin Wan, Peizhao Zhang, Zhicheng Yan, Masayoshi Tomizuka, Joseph Gonzalez, Kurt Keutzer, and Peter Vajda.
\newblock Visual transformers: Token-based image representation and processing for computer vision, 2020.

\bibitem{rombach2022high}
Robin Rombach, Andreas Blattmann, Dominik Lorenz, Patrick Esser, and Bj{\"o}rn Ommer.
\newblock High-resolution image synthesis with latent diffusion models.
\newblock In {\em CVPR}, 2022.

\bibitem{baevski2020wav2vec}
Alexei Baevski, Yuhao Zhou, Abdelrahman Mohamed, and Michael Auli.
\newblock wav2vec 2.0: A framework for self-supervised learning of speech representations.
\newblock {\em Advances in neural information processing systems}, 33:12449--12460, 2020.

\bibitem{whisper}
Alec Radford, Jong~Wook Kim, Tao Xu, Greg Brockman, Christine McLeavey, and Ilya Sutskever.
\newblock Robust speech recognition via large-scale weak supervision.
\newblock In {\em Proc. International Conference on Machine Learning}, pages 28492--28518, Honolulu, USA, July 2023.

\bibitem{brown2020language}
Tom Brown, Benjamin Mann, Nick Ryder, Melanie Subbiah, Jared~D Kaplan, Prafulla Dhariwal, Arvind Neelakantan, Pranav Shyam, Girish Sastry, Amanda Askell, et~al.
\newblock Language models are few-shot learners.
\newblock In {\em NeurIPS}, 2020.

\bibitem{grattafiori2024llama}
Aaron Grattafiori, Abhimanyu Dubey, Abhinav Jauhri, Abhinav Pandey, Abhishek Kadian, Ahmad Al-Dahle, Aiesha Letman, Akhil Mathur, Alan Schelten, Alex Vaughan, et~al.
\newblock The llama 3 herd of models.
\newblock {\em arXiv preprint arXiv:2407.21783}, 2024.

\bibitem{houlsby2019parameter}
Neil Houlsby, Andrei Giurgiu, Stanislaw Jastrzebski, Bruna Morrone, Quentin De~Laroussilhe, Andrea Gesmundo, Mona Attariyan, and Sylvain Gelly.
\newblock Parameter-efficient transfer learning for nlp.
\newblock In {\em ICML}, 2019.

\bibitem{liu2021gpt}
X~Liu, Y~Zheng, Z~Du, M~Ding, Y~Qian, Z~Yang, and J~Tang.
\newblock Gpt understands, too. arxiv preprint arxiv: 210310385.
\newblock 2021.

\bibitem{lester2021power}
Brian Lester, Rami Al-Rfou, and Noah Constant.
\newblock The power of scale for parameter-efficient prompt tuning.
\newblock {\em arXiv preprint arXiv:2104.08691}, 2021.

\bibitem{hu2022lora}
Edward~J Hu, Phillip Wallis, Zeyuan Allen-Zhu, Yuanzhi Li, Shean Wang, Lu~Wang, Weizhu Chen, et~al.
\newblock Lora: Low-rank adaptation of large language models.
\newblock In {\em ICLR}, 2022.

\bibitem{huang2025hira}
Qiushi Huang, Tom Ko, Zhan Zhuang, Lilian Tang, and Yu~Zhang.
\newblock Hira: Parameter-efficient hadamard high-rank adaptation for large language models.
\newblock In {\em The Thirteenth International Conference on Learning Representations}, 2025.

\bibitem{aghajanyan2020intrinsic}
Armen Aghajanyan, Luke Zettlemoyer, and Sonal Gupta.
\newblock Intrinsic dimensionality explains the effectiveness of language model fine-tuning.
\newblock {\em arXiv preprint arXiv:2012.13255}, 2020.

\bibitem{liu2024dora}
Shih-yang Liu, Chien-Yi Wang, Hongxu Yin, Pavlo Molchanov, Yu-Chiang~Frank Wang, Kwang-Ting Cheng, and Min-Hung Chen.
\newblock Dora: Weight-decomposed low-rank adaptation.
\newblock In {\em ICML}, 2024.

\bibitem{li2021prefixtuningoptimizingcontinuousprompts}
Xiang~Lisa Li and Percy Liang.
\newblock Prefix-tuning: Optimizing continuous prompts for generation, 2021.

\bibitem{zaken2022bitfitsimpleparameterefficientfinetuning}
Elad~Ben Zaken, Shauli Ravfogel, and Yoav Goldberg.
\newblock Bitfit: Simple parameter-efficient fine-tuning for transformer-based masked language-models, 2022.

\bibitem{zhang2023adalora}
Qingru Zhang, Minshuo Chen, Alexander Bukharin, Nikos Karampatziakis, Pengcheng He, Yu~Cheng, Weizhu Chen, and Tuo Zhao.
\newblock Adalora: Adaptive budget allocation for parameter-efficient fine-tuning.
\newblock {\em arXiv preprint arXiv:2303.10512}, 2023.

\bibitem{hayou2024loraefficientlowrank}
Soufiane Hayou, Nikhil Ghosh, and Bin Yu.
\newblock Lora+: Efficient low rank adaptation of large models, 2024.

\bibitem{meng2024pissa}
Fanxu Meng, Zhaohui Wang, and Muhan Zhang.
\newblock Pissa: Principal singular values and singular vectors adaptation of large language models.
\newblock {\em arXiv preprint arXiv:2404.02948}, 2024.

\bibitem{kalajdzievski2023rank}
Damjan Kalajdzievski.
\newblock A rank stabilization scaling factor for fine-tuning with lora.
\newblock {\em arXiv preprint arXiv:2312.03732}, 2023.

\bibitem{dettmers2023qlora}
Tim Dettmers, Artidoro Pagnoni, Ari Holtzman, and Luke Zettlemoyer.
\newblock Qlora: Efficient finetuning of quantized llms.
\newblock {\em Advances in neural information processing systems}, 36:10088--10115, 2023.

\bibitem{xu2023qa}
Yuhui Xu, Lingxi Xie, Xiaotao Gu, Xin Chen, Heng Chang, Hengheng Zhang, Zhengsu Chen, Xiaopeng Zhang, and Qi~Tian.
\newblock Qa-lora: Quantization-aware low-rank adaptation of large language models.
\newblock {\em arXiv preprint arXiv:2309.14717}, 2023.

\bibitem{kingma2022autoencodingvariationalbayes}
Diederik~P Kingma and Max Welling.
\newblock Auto-encoding variational bayes, 2022.

\bibitem{locatello2018challenging}
F~Locatello, S~Bauer, M~Lucic, G~R{\"a}tsch, S~Gelly, B~Sch{\"o}lkopf, and O~Bachem.
\newblock Challenging common assumptions in the unsupervised learning of disentangled representations. arxiv preprint arxiv: 1811.12359, 2018.

\bibitem{higgins2017beta}
Irina Higgins, Loic Matthey, Arka Pal, Christopher Burgess, Xavier Glorot, Matthew Botvinick, Shakir Mohamed, and Alexander Lerchner.
\newblock beta-vae: Learning basic visual concepts with a constrained variational framework.
\newblock In {\em International conference on learning representations}, 2017.

\bibitem{hsu2017unsupervised}
Wei-Ning Hsu, Yu~Zhang, and James Glass.
\newblock Unsupervised learning of disentangled and interpretable representations from sequential data.
\newblock {\em Advances in neural information processing systems}, 30, 2017.

\bibitem{kim2019disentanglingfactorising}
Hyunjik Kim and Andriy Mnih.
\newblock Disentangling by factorising, 2019.

\bibitem{chen2019isolatingsourcesdisentanglementvariational}
Ricky T.~Q. Chen, Xuechen Li, Roger Grosse, and David Duvenaud.
\newblock Isolating sources of disentanglement in variational autoencoders, 2019.

\bibitem{kumar2018variationalinferencedisentangledlatent}
Abhishek Kumar, Prasanna Sattigeri, and Avinash Balakrishnan.
\newblock Variational inference of disentangled latent concepts from unlabeled observations, 2018.

\bibitem{burgess2018understandingdisentanglingbetavae}
Christopher~P. Burgess, Irina Higgins, Arka Pal, Loic Matthey, Nick Watters, Guillaume Desjardins, and Alexander Lerchner.
\newblock Understanding disentangling in $\beta$-vae, 2018.

\bibitem{cimpoi2014describing}
Mircea Cimpoi, Subhransu Maji, Iasonas Kokkinos, Sammy Mohamed, and Andrea Vedaldi.
\newblock Describing textures in the wild.
\newblock In {\em CVPR}, 2014.

\bibitem{helber2019eurosat}
Patrick Helber, Benjamin Bischke, Andreas Dengel, and Damian Borth.
\newblock Eurosat: A novel dataset and deep learning benchmark for land use and land cover classification.
\newblock {\em IEEE Journal of Selected Topics in Applied Earth Observations and Remote Sensing}, 12(7):2217--2226, 2019.

\bibitem{houben2013detection}
Sebastian Houben, Johannes Stallkamp, Jan Salmen, Marc Schlipsing, and Christian Igel.
\newblock Detection of traffic signs in real-world images: The german traffic sign detection benchmark.
\newblock In {\em IJCNN}, 2013.

\bibitem{cheng2017remote}
Gong Cheng, Junwei Han, and Xiaoqiang Lu.
\newblock Remote sensing image scene classification: Benchmark and state of the art.
\newblock {\em Proceedings of the IEEE}, 105(10):1865--1883, 2017.

\bibitem{xiao2010sun}
Jianxiong Xiao, James Hays, Krista~A Ehinger, Aude Oliva, and Antonio Torralba.
\newblock Sun database: Large-scale scene recognition from abbey to zoo.
\newblock In {\em CVPR}, 2010.

\bibitem{netzer2011reading}
Yuval Netzer, Tao Wang, Adam Coates, Alessandro Bissacco, Baolin Wu, Andrew~Y Ng, et~al.
\newblock Reading digits in natural images with unsupervised feature learning.
\newblock In {\em NeurIPS workshop}, 2011.

\bibitem{buyukakyuz2024olora}
Kerim B{\"u}y{\"u}kaky{\"u}z.
\newblock Olora: Orthonormal low-rank adaptation of large language models.
\newblock {\em arXiv preprint arXiv:2406.01775}, 2024.

\bibitem{hu2023llm}
Zhiqiang Hu, Lei Wang, Yihuai Lan, Wanyu Xu, Ee-Peng Lim, Lidong Bing, Xing Xu, Soujanya Poria, and Roy Ka-Wei Lee.
\newblock Llm-adapters: An adapter family for parameter-efficient fine-tuning of large language models.
\newblock {\em arXiv preprint arXiv:2304.01933}, 2023.

\bibitem{piqa}
Yonatan Bisk, Rowan Zellers, Ronan~Le Bras, Jianfeng Gao, and Yejin Choi.
\newblock Piqa: Reasoning about physical commonsense in natural language.
\newblock In {\em Thirty-Fourth AAAI Conference on Artificial Intelligence}, 2020.

\bibitem{siqa}
Maarten Sap, Hannah Rashkin, Derek Chen, Ronan LeBras, and Yejin Choi.
\newblock Socialiqa: Commonsense reasoning about social interactions.
\newblock {\em arXiv preprint arXiv:1904.09728}, 2019.

\bibitem{arc}
Peter Clark, Isaac Cowhey, Oren Etzioni, Tushar Khot, Ashish Sabharwal, Carissa Schoenick, and Oyvind Tafjord.
\newblock Think you have solved question answering? try arc, the ai2 reasoning challenge.
\newblock {\em arXiv:1803.05457v1}, 2018.

\bibitem{mihaylov2018can}
Todor Mihaylov, Peter Clark, Tushar Khot, and Ashish Sabharwal.
\newblock Can a suit of armor conduct electricity? a new dataset for open book question answering.
\newblock {\em arXiv preprint arXiv:1809.02789}, 2018.

\bibitem{zellers2019hellaswag}
Rowan Zellers, Ari Holtzman, Yonatan Bisk, Ali Farhadi, and Yejin Choi.
\newblock Hellaswag: Can a machine really finish your sentence?
\newblock {\em arXiv preprint arXiv:1905.07830}, 2019.

\bibitem{winogrande}
Keisuke Sakaguchi, Ronan~Le Bras, Chandra Bhagavatula, and Yejin Choi.
\newblock Winogrande: An adversarial winograd schema challenge at scale.
\newblock {\em Communications of the ACM}, 64(9):99--106, 2021.

\bibitem{wang2019glue}
Alex Wang, Amanpreet Singh, Julian Michael, Felix Hill, Omer Levy, and Samuel~R Bowman.
\newblock Glue: A multi-task benchmark and analysis platform for natural language understanding.
\newblock In {\em ICLR}, 2019.

\bibitem{liu2019roberta}
Yinhan Liu, Myle Ott, Naman Goyal, Jingfei Du, Mandar Joshi, Danqi Chen, Omer Levy, Mike Lewis, Luke Zettlemoyer, and Veselin Stoyanov.
\newblock Roberta: A robustly optimized bert pretraining approach.
\newblock {\em arXiv preprint arXiv:1907.11692}, 2019.

\bibitem{liu-etal-2022-p}
Xiao Liu, Kaixuan Ji, Yicheng Fu, Weng Tam, Zhengxiao Du, Zhilin Yang, and Jie Tang.
\newblock {P}-tuning: Prompt tuning can be comparable to fine-tuning across scales and tasks.
\newblock In Smaranda Muresan, Preslav Nakov, and Aline Villavicencio, editors, {\em Proceedings of the 60th Annual Meeting of the Association for Computational Linguistics (Volume 2: Short Papers)}, pages 61--68, Dublin, Ireland, May 2022. Association for Computational Linguistics.

\bibitem{garofolo1993timit}
John~S Garofolo, Lori~F Lamel, William~M Fisher, David~S Pallett, Nancy~L Dahlgren, Victor Zue, and Jonathan~G Fiscus.
\newblock Timit acoustic-phonetic continuous speech corpus.
\newblock {\em (No Title)}, 1993.

\bibitem{graves2006connectionist}
Alex Graves, Santiago Fern{\'a}ndez, Faustino Gomez, and J{\"u}rgen Schmidhuber.
\newblock Connectionist temporal classification: {L}abelling unsegmented sequence data with recurrent neural networks.
\newblock In {\em Proc. International Conference on Machine learning}, pages 369--376, Pittsburgh, USA, June 2006.

\bibitem{qiu2024complexity}
Guanwen Qiu, Da~Kuang, and Surbhi Goel.
\newblock Complexity matters: Feature learning in the presence of spurious correlations.
\newblock In {\em Proceedings of the 41st International Conference on Machine Learning}, volume 235, pages 41658--41697, 2024.

\bibitem{sreekumar2023spurious}
Gautam Sreekumar and Vishnu~Naresh Boddeti.
\newblock Spurious correlations and where to find them.
\newblock {\em arXiv preprint arXiv:2308.11043}, 2023.

\bibitem{pmlr-v119-sagawa20a}
Shiori Sagawa, Aditi Raghunathan, Pang~Wei Koh, and Percy Liang.
\newblock An investigation of why overparameterization exacerbates spurious correlations.
\newblock In Hal~Daumé III and Aarti Singh, editors, {\em Proceedings of the 37th International Conference on Machine Learning}, volume 119 of {\em Proceedings of Machine Learning Research}, pages 8346--8356. PMLR, 13--18 Jul 2020.

\bibitem{cui2024ameliorate}
Justin Cui, Ruochen Wang, Yuanhao Xiong, and Cho-Jui Hsieh.
\newblock Ameliorate spurious correlations in dataset condensation.
\newblock In {\em Proceedings of the 41st International Conference on Machine Learning}, volume 235, pages 9696--9721, 2024.

\bibitem{pmlr-v119-srivastava20a}
Megha Srivastava, Tatsunori Hashimoto, and Percy Liang.
\newblock Robustness to spurious correlations via human annotations.
\newblock In Hal~Daumé III and Aarti Singh, editors, {\em Proceedings of the 37th International Conference on Machine Learning}, volume 119 of {\em Proceedings of Machine Learning Research}, pages 9109--9119. PMLR, 13--18 Jul 2020.

\bibitem{koh2021wilds}
Pang~Wei Koh, Shiori Sagawa, Henrik Marklund, Sang~Michael Xie, Marvin Zhang, Akshay Balsubramani, Weihua Hu, Michihiro Yasunaga, Richard~Lanas Phillips, Irena Gao, et~al.
\newblock Wilds: A benchmark of in-the-wild distribution shifts.
\newblock In {\em International conference on machine learning}, pages 5637--5664. PMLR, 2021.

\bibitem{joshi2025challenges}
Siddharth Joshi, Yu~Yang, Yihao Xue, Wenhan Yang, and Baharan Mirzasoleiman.
\newblock Challenges and opportunities in improving worst-group generalization in presence of spurious features, 2025.

\bibitem{deng2009imagenet}
Jia Deng, Wei Dong, Richard Socher, Li-Jia Li, Kai Li, and Li~Fei-Fei.
\newblock Imagenet: A large-scale hierarchical image database.
\newblock In {\em 2009 IEEE conference on computer vision and pattern recognition}, pages 248--255. Ieee, 2009.

\bibitem{sagawa2020distributionally}
Shiori Sagawa*, Pang~Wei Koh*, Tatsunori~B. Hashimoto, and Percy Liang.
\newblock Distributionally robust neural networks.
\newblock In {\em International Conference on Learning Representations}, 2020.

\bibitem{creager2021environment}
Elliot Creager, J{\"o}rn-Henrik Jacobsen, and Richard Zemel.
\newblock Environment inference for invariant learning.
\newblock In {\em International Conference on Machine Learning}, pages 2189--2200. PMLR, 2021.

\bibitem{liu2021just}
Evan~Z Liu, Behzad Haghgoo, Annie~S Chen, Aditi Raghunathan, Pang~Wei Koh, Shiori Sagawa, Percy Liang, and Chelsea Finn.
\newblock Just train twice: Improving group robustness without training group information.
\newblock In {\em International Conference on Machine Learning}, pages 6781--6792. PMLR, 2021.

\end{thebibliography}

\newpage
\appendix

\section{Variational Auto-Encoder}

\subsection{VAE Objective Derivation}
\label{appendix:elbo_derivation}

We derive the Evidence Lower Bound (ELBO) by starting from the marginal log-likelihood:
\begin{align*}
\log p_{\theta}(\vx) 
&= \mathbb{E}_{\vz \sim q_\phi(\vz |\vx)} \left[ \log p_{\theta}(\vx) \right] \\
&= \mathbb{E}_{\vz \sim q_\phi(\vz |\vx)} \left[ \log \left( \frac{p_{\theta}(\vx |\vz) p(\vz)}{p_{\theta}(\vz |\vx)} \right) \right] \\
&= \mathbb{E}_{\vz \sim q_\phi(\vz |\vx)} \left[ \log p_{\theta}(\vx |\vz) + \log \frac{p(\vz)}{q_\phi(\vz |\vx)} + \log \frac{q_\phi(\vz |\vx)}{p_{\theta}(\vz |\vx)} \right] \\
&= \mathbb{E}_{\vz \sim q_\phi(\vz |\vx)} \left[ \log p_{\theta}(\vx |\vz) \right] - D_{\mathrm{KL}}\left(q_\phi(\vz |\vx) \,\|\, p(\vz)\right) + D_{\mathrm{KL}}\left(q_\phi(\vz |\vx) \,\|\, p_{\theta}(\vz |\vx)\right).
\end{align*}

The last term is always non-negative, which justifies interpreting the remaining two terms as a lower bound, i.e.,
\[
\log p_{\theta}(\vx) \geq \mathcal{L}_{\theta, \phi}^{\text{VAE}}(\vx),
\]
with
\[
\mathcal{L}_{\theta, \phi}^{\text{VAE}}(\vx) = \mathbb{E}_{\vz \sim q_\phi(\vz |\vx)} \left[ \log p_\theta(\vx |\vz) \right] - D_{\mathrm{KL}}\left(q_\phi(\vz |\vx) \,\|\, p(\vz)\right).
\]

\subsection{VAE2LAT:~VAE with 2 Latent Variables Objective Derivation}
\label{appendix:elbo2_derivation}

We simply start from the ELBO previously derived with two variables, i.e.,

\[
\mathcal{L}_{\theta, \phi}^{\text{VAE2LAT}}(\vx) = \mathbb{E}_{\vz \sim q_\phi(\vz_1,\vz_2 | \vx)} \left[ \log p_\theta(\vx | \vz_1,\vz_2) \right] - D_{\mathrm{KL}}\left(q_\phi(\vz_1,\vz_2 | \vx) \,\|\, p(\vz_1,\vz_2)\right).
\]

Applying the independence assumption, we obtain

\begin{align}
\mathcal{L}_{\theta, \phi}^{\text{VAE2LAT}}(\vx) = \mathbb{E}_{\substack{\vz_1 \sim q_{\phi_1}(\vz_1 | \vx) \\ \vz_2 \sim q_{\phi_2}(\vz_2 | \vx)}} \left[\log p_\theta(\vx | \vz_1, \vz_2)\right] 
- D_{\mathrm{KL}}\left(q_{\phi_1}(\vz_1 | \vx) \,\|\, p_1(\vz_1)\right) - D_{\mathrm{KL}}\left(q_{\phi_2}(\vz_2 | \vx) \,\|\, p_2(\vz_2)\right).
\end{align}

\subsection{$\beta$-VAE2LAT:~A $\beta$-VAE with 2 latents variables }
\label{appendix:elbo3_derivation}

The loss of $\beta$-VAE2LAT, i.e., a straightforward extension of $\beta$-VAE to two latent variables is given by studies~\ref{sec:res_ablations}.
\begin{align}
\label{eq:betavae2lat}
\mathcal{L}_{\theta, \phi}^{\vb-\text{VAE2LAT}}(\vx) = \mathbb{E}_{\substack{\vz_1 \sim q_{\phi_1}(\vz_1 | \vx) \\ \vz_2 \sim q_{\phi_2}(\vz_2 | \vx)}} \left[\log p_\theta(\vx | \vz_1, \vz_2)\right] 
- \beta D_{\mathrm{KL}}\left(q_{\phi_1}(\vz_1 | \vx) \,\|\, p_1(\vz_1)\right) - \beta D_{\mathrm{KL}}\left(q_{\phi_2}(\vz_2 | \vx) \,\|\, p_2(\vz_2)\right).
\end{align}
This formulation is used in the ablation studies in Section~\ref{sec:res_ablations}.

\section{FVAE}

\subsection{Bounding the Discrepancy Term \(\Delta\) via the 2-Wasserstein Distance}
\label{app:delta_bound}

We bound the discrepancy
\[
\Delta
\;=\;
\mathbb{E}_{\vz \sim q_{\phi_2}}\!\left[\log p_1(\vz)\right]
\;-\;
\mathbb{E}_{\vz \sim q_{\phi_1}}\!\left[\log p_1(\vz)\right],
\]
assuming only that the log-prior \(f(z)=\log p_1(z)\) is \(C^2\) with a globally
bounded Hessian:
\[
\;
\|\nabla^2 f(z)\|_\mathrm{op}\;\le\;L\quad\forall z\in\mathbb{R}^d .
\;
\tag{H}
\]

\vspace{.5em}
\paragraph{Step 1 – Second-order Taylor control.}
For any two points \(z_1,z_2\), Taylor’s formula with integral remainder gives
\[
f(z_2)-f(z_1)
\;=\;
\langle\nabla f(z_1),\,z_2-z_1\rangle
\;+\;
(z_2-z_1)^{\!\top}\!\!
\left(\int_0^1 (1-s)\,\nabla^2\!f\bigl(z_1+s\,(z_2-z_1)\bigr)ds\right)
(z_2-z_1).
\]
Bounding the remainder using assumption (H) gives the point-wise inequality
\[
\bigl|f(z_2)-f(z_1)-\langle\nabla f(z_1),z_2-z_1\rangle\bigr|
\;\le\;
\frac{L}{2}\,\|z_2-z_1\|^2 .
\tag{A}
\]

\vspace{.5em}
\paragraph{Step 2 – Integrate over a coupling.}
Let \(\gamma\in\Pi(q_{\phi_1},q_{\phi_2})\) be \emph{any} coupling of the two
distributions, and write \((\vz_1,\vz_2)\sim\gamma\), \(d=\vz_2-\vz_1\).
Taking expectations in (A), applying the triangle inequality, and then Cauchy–Schwarz to the linear term,
\[
\bigl|\Delta\bigr|
\;\le\;
\frac{L}{2}\;\mathbb{E}_\gamma\|d\|^2
\;+\;
\underbrace{\bigl|\mathbb{E}_\gamma\langle\nabla f(\vz_1),d\rangle\bigr|}
    _{\text{``linear term expectation''}}
\;\le\;
\frac{L}{2}\,\mathbb{E}_\gamma\|d\|^2
\;+\;
\sqrt{\mathbb{E}_{q_{\phi_1}}\!\|\nabla f\|^2}\;
    \sqrt{\mathbb{E}_\gamma\|d\|^2}.
\]
Now minimise the rightmost expression over \(\gamma\). Since the function $g(x) = \frac{L}{2}x^2 + Cx$ (for $C=\sqrt{\mathbb{E}_{q_{\phi_1}}\!\|\nabla f\|^2} \ge 0$) is non-decreasing for $x = \sqrt{\mathbb{E}_\gamma\|d\|^2} \ge 0$, the infimum is attained when $\mathbb{E}_\gamma\|d\|^2$ is minimized. The infimum of
\(\mathbb{E}_\gamma\|d\|^2\) is \(\mathcal{W}_2^2(q_{\phi_1},q_{\phi_2})\). Hence~:
\[
\;
\bigl|\Delta\bigr|
\;\le\;
\frac{L}{2}\,\mathcal{W}_2^2(q_{\phi_1},q_{\phi_2})
\;+\;
\sqrt{\mathbb{E}_{q_{\phi_1}}\!\bigl\|\nabla \log p_1(\vz)\bigr\|^2}\;
    \mathcal{W}_2(q_{\phi_1},q_{\phi_2}).
\;
\]

\vspace{.5em}
\paragraph{Step 3 – Specialization to Gaussian case.}
Assume \( p_1 = \mathcal{N}(0, I) \) and \( q_{\phi_1} = \mathcal{N}(\boldsymbol{\mu}, \operatorname{diag}(\boldsymbol{\sigma}^2)) \). Then the gradient becomes \( \nabla \log p_1(\vz) = -\vz \), and the expectation simplifies as:
\[
\mathbb{E}_{q_{\phi_1}}\!\left\| \nabla \log p_1(\vz) \right\|^2 = \mathbb{E}_{q_{\phi_1}}\!\left[ \|\vz\|^2 \right] = \sum_j \mu_j^2 + \sigma_j^2.
\]
Therefore, the bound becomes:
\[
\bigl|\Delta\bigr|
\;\le\;
\frac{1}{2}\,\mathcal{W}_2^2(q_{\phi_1},q_{\phi_2})
\;+\;
\sqrt{ \sum_j \mu_j^2 + \sigma_j^2 } \cdot \mathcal{W}_2(q_{\phi_1},q_{\phi_2}).
\]
Since \( q_{\phi_1} \) is trained to approximate \( p_1 \), the square-root term is typically bounded in practice. Hence, both terms contribute to increasing \( \mathcal{W}_2(q_{\phi_1}, q_{\phi_2}) \), and the discrepancy \( \Delta \) serves as an effective Wasserstein repulsion.

\section{Dataset Details}
\label{appendix:dataset_details}

\subsection{Spurious Correlation Experiments}
\label{appendix:dataset_details_spurious}
\begin{table}[H]
\centering
\caption{Statistics of the datasets used in the spurious experiment.}
\begin{tabular}{l|cccc}
\hline
\textbf{Dataset} & SpuCoAnimals & Waterbirds & CelebA     \\ \hline
\# Classes       & 4            & 2          & 2          \\
\# Groups        & 8            & 4          & 4          \\ \hline
Train            & 42000        & 4795       & 162770     \\
Validation             & 2100         & 1199       & 19867\\
Test             & 4000         & 5794       & 19962      \\ \hline
Class Ratio      & 25:25:25:25  & 76.8:23.2  & 85:15      \\
\end{tabular}%
\label{tab:spuco_datasets}
\end{table}

\section{Hyperparameters}
\label{appendix:hparams_details}
This section details the hyperparameters used for the experiments presented in the main paper. For all LoRA-based methods, including FVAE-LoRA, the LoRA rank ($r$) was set to 16, and LoRA was applied to the query and key matrices of the attention layers.
The latent dimension of $\vz_1$ in FVAE-LoRA corresponds to this LoRA rank.

\subsection{Image Experiments}
\label{appendix:hparams_details_image}
The following hyperparameters were used for fine-tuning ViT-B/16 on DTD, EuroSAT, GTSRB, RESISC45, SUN397, and SVHN datasets.
\begin{table}[H]
\centering
\caption{Hyperparameters for Image Classification tasks using ViT-B/16.}
\label{tab:hparams_image}
\small 
\begin{tabular}{@{}ll@{}}
\toprule
Parameter                     & Value / Setting                 \\
\midrule
\multicolumn{2}{@{}l}{\textit{General Training Parameters}} \\
Optimizer                     & AdamW                         \\
Learning Rate                 & $5 \times 10^{-3}$           \\
LR Scheduler                  & Linear              \\
Warmup Ratio                 & 0.1                             \\
Batch Size                    & 32                            \\
Number of Epochs              & 30                            \\
Weight Decay                  & $0.01$                        \\
Seeds                          & 1, 2, 42                            \\
\midrule
\multicolumn{2}{@{}l}{\textit{LoRA Parameters}} \\
LoRA Rank ($r$)               & 16                            \\
LoRA Dropout                  & 0.1                           \\
\midrule
\multicolumn{2}{@{}l}{\textit{FVAE-LoRA Specific Parameters}} \\
Latent Dim. $\vz_1$           & 16 (same as LoRA rank)        \\
Latent Dim. $\vz_2$           & 16                            \\
FVAE $q_{\phi_i}(\vz_i|\vx)$ Enc. Arch. & $\vx \xrightarrow{\text{Linear}}$ dim($\vz_i$) $\xrightarrow{\text{ReLU}} \text{HiddenState}_{\vz_i} \xrightarrow{\text{Linear}} (\boldsymbol{\mu}_{\vz_i}, \log\boldsymbol{\sigma}^2_{\vz_i})$ \\
FVAE $p_{\theta}(\vx|\vz_1, \vz_2)$ Dec. Arch. & Concat($\vz_1, \vz_2$) $\xrightarrow{\text{Linear}}$ $H_D=128 \xrightarrow{\text{ReLU}}$ Linear $\xrightarrow{} \hat{\vx}$ (Input Dim) \\
Prior $p_1(\vz_1)$            & $\mathcal{N}(0, I)$           \\
Prior $p_2(\vz_2)$            & $\mathcal{N}(1.5, I)$         \\
$\vl$ (Eq.~\ref{eq:to_refer_downstream})           & $1 \times 10^{-3}$ \\
ELBO Coeff. $\alpha$ (Reconstr.) & $1$                        \\
ELBO Coeff. $\beta$ (KL $q_1||p_1$) & $1$ or $10$                         \\
ELBO Coeff. $\delta$ & $1$                        \\
\bottomrule
\end{tabular}
\end{table}

\subsection{Text Experiments}
\label{appendix:hparams_details_text}
\begin{table}[H]
\centering
\caption{Hyperparameters for Commonsense Reasoning using Llama-3-8B.}
\label{tab:hparams_text_llama}
\small
\begin{tabular}{@{}ll@{}}
\toprule
Parameter                     & Value / Setting                 \\
\midrule
\multicolumn{2}{@{}l}{\textit{General Training Parameters}} \\
Optimizer                     & AdamW                         \\
Learning Rate                 & $1 \times 10^{-3}$ (LoRA, HiRA),  $3 \times 10^{-4}$ (FVAE-LoRA)            \\
LR Scheduler                  & Linear      \\
Warmup Steps                  & 100                          \\
Batch Size       & 8                             \\
Gradient Accumulation Steps   & 4                             \\
Number of Epochs              & 3                             \\
Weight Decay                  & $0.0$                         \\
Seed                          & 42                            \\
\midrule
\multicolumn{2}{@{}l}{\textit{LoRA Parameters}} \\
LoRA Rank ($r$)               & 16                            \\
LoRA Dropout                  & 0.1                          \\
Target Modules                & q\_proj, k\_proj            \\ 
\midrule
\multicolumn{2}{@{}l}{\textit{FVAE-LoRA Specific Parameters}} \\
Latent Dim. $\vz_1$           & 16                            \\
Latent Dim. $\vz_2$           & 16                            \\
FVAE $q_{\phi_i}(\vz_i|\vx)$ Enc. Arch. & $\vx \xrightarrow{\text{Linear}}$ dim($\vz_i$) $\xrightarrow{\text{ReLU}} \text{HiddenState}_{\vz_i} \xrightarrow{\text{Linear}} (\boldsymbol{\mu}_{\vz_i}, \log\boldsymbol{\sigma}^2_{\vz_i})$ \\
FVAE $p_{\theta}(\vx|\vz_1, \vz_2)$ Dec. Arch. & Concat($\vz_1, \vz_2$) $\xrightarrow{\text{Linear}}$ $H_D=128 \xrightarrow{\text{ReLU}}$ Linear $\xrightarrow{} \hat{\vx}$ (Input Dim) \\
Prior $p_1(\vz_1)$            & $\mathcal{N}(0, I)$           \\
Prior $p_2(\vz_2)$            & $\mathcal{N}(1.5, I)$         \\
$\vl$ (Eq.~\ref{eq:to_refer_downstream})           & $1 \times 10^{-4}$ \\
ELBO Coeff. $\alpha$ (Reconstr.) & $1$                        \\
ELBO Coeff. $\beta$ (KL $q_1||p_1$) & $1$ or $10$                         \\
ELBO Coeff. $\delta$ & $1$                        \\
\bottomrule
\end{tabular}
\end{table}

\begin{table}[H]
\centering
\caption{Hyperparameters for GLUE Benchmark tasks using RoBERTa-base.}
\label{tab:hparams_text_roberta}
\small
\begin{tabular}{@{}ll@{}}
\toprule
Parameter                     & Value / Setting                 \\
\midrule
\multicolumn{2}{@{}l}{\textit{General Training Parameters}} \\
Optimizer                     & AdamW                         \\
Learning Rate                 & $3 \times 10^{-4}$ \\
LR Scheduler                  & Linear      \\
Warmup Ratio                  & 0.06        \\
Batch Size                    & 32     \\
Number of Epochs              & 30 \\
Seed                          & 1, 2, 42                            \\
\midrule
\multicolumn{2}{@{}l}{\textit{LoRA Parameters}} \\
LoRA Rank ($r$)               & 16                            \\
LoRA Dropout                  & 0.1                           \\
\midrule
\multicolumn{2}{@{}l}{\textit{FVAE-LoRA Specific Parameters}} \\
Latent Dim. $\vz_1$           & 16                            \\
Latent Dim. $\vz_2$           & 16                            \\
FVAE $q_{\phi_i}(\vz_i|\vx)$ Enc. Arch. & $\vx \xrightarrow{\text{Linear}}$ dim($\vz_i$) $\xrightarrow{\text{ReLU}} \text{HiddenState}_{\vz_i} \xrightarrow{\text{Linear}} (\boldsymbol{\mu}_{\vz_i}, \log\boldsymbol{\sigma}^2_{\vz_i})$ \\
FVAE $p_{\theta}(\vx|\vz_1, \vz_2)$ Dec. Arch. & Concat($\vz_1, \vz_2$) $\xrightarrow{\text{Linear}}$ $H_D=128 \xrightarrow{\text{ReLU}}$ Linear $\xrightarrow{} \hat{\vx}$ (Input Dim) \\
Prior $p_1(\vz_1)$            & $\mathcal{N}(0, I)$           \\
Prior $p_2(\vz_2)$            & $\mathcal{N}(1.5, I)$         \\
$\vl$ (Eq.~\ref{eq:to_refer_downstream})           & $1 \times 10^{-3}$ or $1 \times 10^{-4}$ \\
ELBO Coeff. $\alpha$ (Reconstr.) & $0.1$ or $1$                        \\
ELBO Coeff. $\beta$ (KL $q_1||p_1$) & $1$ or $10$                         \\
ELBO Coeff. $\delta$ & $1$                        \\
\bottomrule
\end{tabular}
\end{table}

\subsection{Audio Experiments}
\label{appendix:hparams_details_audio}
The following hyperparameters were used for fine-tuning Wav2Vec2-Large on the TIMIT dataset.

\begin{table}[H]
\centering
\caption{Hyperparameters for ASR on TIMIT using Wav2Vec2-Large.}
\label{tab:hparams_audio}
\small
\begin{tabular}{@{}ll@{}}
\toprule
Parameter                     & Value / Setting                 \\
\midrule
\multicolumn{2}{@{}l}{\textit{General Training Parameters}} \\
Optimizer                     & AdamW                         \\
Learning Rate                 & $5 \times 10^{-5}$ (Full FT), $5 \times 10^{-4}$ (LoRA and FVAE-LoRA)            \\
LR Scheduler                  & Linear   \\
Warmup Steps                  & 500                           \\
Batch Size                    & 32                             \\
Number of Epochs              & 30                            \\
Weight Decay                  & $0.005$                       \\
CTC Loss Reduction            & Sum                          \\
\midrule
\multicolumn{2}{@{}l}{\textit{LoRA Parameters (Standard LoRA \& FVAE-LoRA's LoRA part)}} \\
LoRA Rank ($r$)               & 16                            \\
LoRA Dropout                  & 0.1                           \\
\midrule
\multicolumn{2}{@{}l}{\textit{FVAE-LoRA Specific Parameters}} \\
Latent Dim. $\vz_1$           & 16                            \\
Latent Dim. $\vz_2$           & 16                            \\
FVAE $q_{\phi_i}(\vz_i|\vx)$ Enc. Arch. & $\vx \xrightarrow{\text{Linear}}$ dim($\vz_i$) $\xrightarrow{\text{ReLU}} \text{HiddenState}_{\vz_i} \xrightarrow{\text{Linear}} (\boldsymbol{\mu}_{\vz_i}, \log\boldsymbol{\sigma}^2_{\vz_i})$ \\
FVAE $p_{\theta}(\vx|\vz_1, \vz_2)$ Dec. Arch. & Concat($\vz_1, \vz_2$) $\xrightarrow{\text{Linear}}$ $H_D=128 \xrightarrow{\text{ReLU}}$ Linear $\xrightarrow{} \hat{\vx}$ (Input Dim) \\
Prior $p_1(\vz_1)$            & $\mathcal{N}(0, I)$           \\
Prior $p_2(\vz_2)$            & $\mathcal{N}(1.5, I)$         \\
$\vl$ (Eq.~\ref{eq:to_refer_downstream})           & $1 \times 10^{-3}$ \\
ELBO Coeff. $\alpha$ (Reconstr.) & $1$                        \\
ELBO Coeff. $\beta$ (KL $q_1||p_1$) & $1$                         \\
ELBO Coeff. $\delta$ & $1$                        \\
\bottomrule
\end{tabular}
\end{table}

\subsection{Spurious Correlation Experiments}
\label{appendix:hparams_details_spurious}
These experiments (Waterbirds, CelebA, Animals) used ViT-B/16 as the backbone. Base training and LoRA parameters are similar to those in Section~\ref{appendix:hparams_details_image}, with specific FVAE-LoRA coefficients tuned for robustness.

\begin{table}[H]
\centering
\caption{Key FVAE-LoRA Hyperparameters for Spurious Correlation tasks (ViT-B/16).}
\label{tab:hparams_spurious}
\small
\begin{tabular}{@{}ll@{}}
\toprule
Parameter                     & Value / Setting                 \\
\midrule
\multicolumn{2}{@{}l}{\textit{General \& LoRA Parameters}} \\
\multicolumn{2}{@{}l}{See Table~\ref{tab:hparams_image} for most general and LoRA parameters.} \\
Batch Size                    & 128     \\
Number of Epochs              & 30      \\
\midrule
\multicolumn{2}{@{}l}{\textit{FVAE-LoRA Specific Parameters}} \\
Latent Dim. $\vz_1$           & 16                            \\
Latent Dim. $\vz_2$           & 16                            \\
FVAE Architecture              & Similar to Table~\ref{tab:hparams_image} \\
Prior $p_1(\vz_1)$            & $\mathcal{N}(0, I)$           \\
Prior $p_2(\vz_2)$            & $\mathcal{N}(1.5, I)$         \\
$\vl$ (Eq.~\ref{eq:to_refer_downstream})           & $1 \times 10^{-3}$ or $1 \times 10^{-4}$ \\
ELBO Coeff. $\alpha$ (Reconstr.) & $0.1$ or $1$                        \\
ELBO Coeff. $\beta$ (KL $q_1||p_1$) & $1$ or $10$                         \\
ELBO Coeff. $\delta$ & $1$                        \\
\bottomrule
\end{tabular}
\end{table}

\section{Early Attempts at Latent Space Factorization}
\label{appendix:early_attempts}
\CR{
The most straightforward way to enforce repulsion between $\vz_1$ and $\vz_2$ in the two-variable ELBO (see Eq. \ref{eq:elbonaive}) would be to augment the ELBO with the terms:
\begin{align}
\label{eq:early_attempts}
+ D_{\mathrm{KL}}\left(q_{\phi_2}\,\|\, p_1\right) + D_{\mathrm{KL}}\left(q_{\phi_1}\,\|\, p_2\right).
\end{align}
However, early experiments with this approach yielded poor results, in fact, performance was worse than with LoRA. From a theoretical standpoint, adding such terms to the two-variable ELBO effectively cancels out $q_{\phi_2}$ and $q_{\phi_1}$ from the objective, leading instead to a direct repulsion between the priors $p_1$ and $p_2$, which is not desirable. Other similar approaches such as directly repelling $q_{\phi_1}$ and $q_{\phi_2}$ suffered from the same issue. We found that all overly symmetric and direct formulations, including two-term symmetric variants, were ultimately unfruitful. To avoid this cancellation effect, we instead propose an indirect way to introduce repulsion between $\vz_1$ and $\vz_2$ by introducing a cross-term between the parametric encoder $q_{\phi_2}$ and the latent distribution $p_1$. This solution is theoretically grounded: we show that it induces a geometric separation, measured through a Wasserstein upper bound, between the two encoders $q_{\phi_1}$ and $q_{\phi_2}$. It is also supported by experimental results, outperforming LoRA across all tested modalities.
}

\section{Additional Insights in FVAE-LoRA}
\label{appendix:additional_insights}
\CR{
Regarding the objective in Eq. \ref{eq:elbo-fvae}, our proposed loss is a novel objective derived from and inspired by the Evidence Lower Bound, but it is not a strict lower bound on the marginal log-likelihood $\log p(\vx)$. By introducing the repulsive regularization term $\Gamma$, we modify the standard ELBO to enforce factorization between the latent spaces. This term is essential for the method's success, but it means the objective no longer serves as a formal lower bound on the data log-likelihood in the traditional VAE sense.}

\CR{FVAE-LoRA intentionally sacrifices static weight merging to enable a more powerful dynamic, input-dependent adaptation. By computing the adaptation specifically for each input $\vx$, our model learns more robust and fine-grained representations. We believe this dynamic mechanism is the key to its performance edge, a capability validated by our strong results on the spurious correlation benchmarks. This trade-off is therefore central to achieving the higher performance and robustness we demonstrate.}

\CR{Note that simply reducing the rank of LoRA is a simple and efficient form of regularization, however It compresses all information flowing through the adapter, without distinguishing between features that are useful, irrelevant, or even detrimental to the downstream task.
Our hypothesis is that large foundation models, pretrained on vast and general datasets, contain rich and entangled set of features. For any specific downstream task, some features are highly relevant (the "signal"), some are irrelevant but harmless, and some are actively harmful. The most prominent example of these detrimental features are spurious correlations (e.g., a water background being correlated with a "waterbird" label). A standard fine-tuning process, which optimizes a task-specific loss, may still latch onto these spurious features because they are prevalent in the training data and help minimize the training loss. This leads to poor generalization on data where that correlation is broken.
This is why FVAE-LoRA is designed to be a more intelligent filter. Its goal is not just to compress, but to actively separate and isolate these different types of information. By using two latent spaces ($\vz_1$ and $\vz_2$) and our novel factorization objective, we encourage the model to encode task-salient, causal information in $\vz_1$ while relegating the residual, non-essential, or spurious information to $\vz_2$.
The most direct validation of this rationale is in our spurious correlation experiments (Section \ref{sec:exp_spurious}). These results show that FVAE-LoRA is significantly more robust to misleading features than standard LoRA, confirming that it successfully learns to rely on the core features isolated within $\vz_1$. This ability to "denoise" the adaptation is why it ultimately achieves better and more reliable performance.}

\section{A Practical Guide on Hyperparamters Selection}
\label{appendix:hparam_guide}
\CR{The factorization in FVAE-LoRA is governed by a subtle equilibrium between reconstruction and regularization, enforced by our ELBO objective. The key hyperparameters, $\beta$ and $\delta$, control this balance.}

\CR{\noindent \textbf{$\beta$ and the Task-Salient Space ($\vz_1$).} \quad The $\beta$ parameter controls the KL divergence on $\vz_1$, our task-salient latent space. Its role is critical, as it enforces a structured and efficient representation of the task-salient features. To understand its impact, we experimented with a wide range of values.}

\CR{A significantly lower value, such as $\beta = 0.1$, led to a drastic degradation in performance across all tasks. This is because a near-zero $\beta$ effectively removes the KL divergence term, freeing the encoder for $\vz_1$ to learn an unconstrained and arbitrarily complex representation. This removes the crucial pressure for the learned posterior $q_{\phi_1}(\vz_1 | \vx)$ to align with the prior $p_1(\vz_1)$, leading to overfitting and a loss of generalization. This result is not merely a poor tuning choice; it is critical evidence that enforcing this prior alignment is essential for learning a robust and meaningful task-salient space.}

\CR{Conversely, we explored a much higher value of $\beta = 100$. While this yielded marginal improvements over $\beta = 10$ on some specific tasks, the gains were not significant enough to justify such a strong constraint. An overly large $\beta$ can create an information bottleneck, punishing the model so heavily for deviating from the prior that it struggles to encode sufficient task-specific information in $\vz_1$.}

\CR{This evidence from both extremes reveals a necessary balance. The optimal values, which we found to be in the range of $1$ to $10$, are large enough to enforce a structured, regularized space but not so large as to prevent the learning of useful features.}

\CR{\noindent \textbf{$\delta$ and Latent Space Separation.} \quad The $\delta$ parameter controls the strength of our repulsive regularizer, $\Gamma$, which is the primary mechanism for enforcing factorization between the task-salient space ($\vz_1$) and the residual space ($\vz_2$). Our empirical results consistently show that $\delta = 1$ provides sufficient repulsive force to achieve this separation effectively, as was demonstrated in the spurious correlation experiments. We recommend $\delta = 1$ as a robust and generally optimal default.}

\CR{For practical application, users can start with $\delta = 1$ and tune $\beta$ (typically between $1$ and $10$) to adjust the regularization on the learned task-salient features. This is a crucial point for the practical application of our method.}

\CR{
\section{Computational Cost Analysis}
\label{appendix:wall_time}
Our empirical results on image classification tasks show that the training time for FVAE-LoRA is approximately 30\% higher than that of the strong DoRA baseline. This increase is primarily due to the additional forward and backward pass through the VAE's decoder during the training phase. However, the inference-time overhead is significantly lower because only the lightweight $\vz_1$ encoder is used.
}


\CR{
\section{Future Work}
\label{appendix:future}
A particularly exciting avenue for future work lies in exploiting the inherent generative capabilities of the FVAE framework. Key directions will include exploiting the generative capabilities of the FVAE decoder for principled data augmentation, applying our latent factorization principle to other PEFT methods beyond LoRA, exploring approximate high-rank adaptation methods like HiRA, and exploring architectural enhancements such as allocating adaptive parameter budgets or different latent space ranks to different layers.
}

\section{Limitations}
\label{appendix:limitations}
While FVAE-LoRA demonstrates promising results across diverse modalities, several limitations of the current work remain. First, the LoRA rank is fixed to a value of 16 across all experiments. Although this ensures consistent parameter budgets, it may not represent the optimal configuration for each task or domain, potentially limiting performance. Second, FVAE-LoRA and all baselines are applied only to the query and key matrices of the transformer models. This restricted application may overlook potential gains from adapting other components such as value matrices or feedforward layers.

Furthermore, detailed hyperparameter settings and VAE-specific architectural choices are provided in the appendix. This separation may hinder reproducibility, for readers interested in extending the approach. Nevertheless, the code will be open-sourced after publication.

In addition, while modality-specific baselines are chosen with the goal of providing meaningful comparisons, we do not evaluate against stronger non-LoRA or non-factorized alternatives, which may offer a more comprehensive picture of relative performance. Further, the paper does not report the computational cost of training or inference, which is important for assessing the practical deployment potential of the method, especially in resource-constrained environments.

Finally, a practical limitation of FVAE-LoRA is that its adapter weights cannot be merged back into the original base model after training. This stands in contrast to some LoRA-based methods that allow for such weight merging, which can simplify inference or reduce model complexity at deployment time.

\section{Broader Impacts}
\label{appendix:broader_impacts}
FVAE-LoRA advances parameter-efficient fine-tuning by enabling a factorized latent representation.
Potential positive impacts include more effective and robust model adaptation across modalities, potentially leading to improved performance, resource efficiency, and more reliable AI systems, especially in handling spurious correlations, as demonstrated in our experiments. This can also enhance accessibility to powerful AI capabilities for a wider range of researchers and developers.

However, techniques that improve the adaptability of large foundation models also carry inherent risks. Easier and more effective fine-tuning could lower barriers for misuse in sensitive areas such as the generation of sophisticated disinformation or the development of enhanced surveillance tools. While FVAE-LoRA aims to disentangle task-salient information, it does not inherently mitigate biases that may be present in the original pre-trained models or the fine-tuning data. Indeed, such biases could potentially be concentrated or even amplified within the task-salient latent space if not proactively identified and addressed.

\newpage
\section*{NeurIPS Paper Checklist}

\begin{enumerate}

\item {\bf Claims}
    \item[] Question: Do the main claims made in the abstract and introduction accurately reflect the paper's contributions and scope?
    \item[] Answer: \answerYes{} 
    \item[] Justification: The main claims in the abstract and introduction accurately reflect the paper's contribution and scope.
    \item[] Guidelines:
    \begin{itemize}
        \item The answer NA means that the abstract and introduction do not include the claims made in the paper.
        \item The abstract and/or introduction should clearly state the claims made, including the contributions made in the paper and important assumptions and limitations. A No or NA answer to this question will not be perceived well by the reviewers. 
        \item The claims made should match theoretical and experimental results, and reflect how much the results can be expected to generalize to other settings. 
        \item It is fine to include aspirational goals as motivation as long as it is clear that these goals are not attained by the paper. 
    \end{itemize}

\item {\bf Limitations}
    \item[] Question: Does the paper discuss the limitations of the work performed by the authors?
    \item[] Answer: \answerYes{} 
    \item[] Justification: The authors provide sufficient information regarding limitations of the current work in the appendix \ref{appendix:limitations}.
    \item[] Guidelines:
    \begin{itemize}
        \item The answer NA means that the paper has no limitation while the answer No means that the paper has limitations, but those are not discussed in the paper. 
        \item The authors are encouraged to create a separate "Limitations" section in their paper.
        \item The paper should point out any strong assumptions and how robust the results are to violations of these assumptions (e.g., independence assumptions, noiseless settings, model well-specification, asymptotic approximations only holding locally). The authors should reflect on how these assumptions might be violated in practice and what the implications would be.
        \item The authors should reflect on the scope of the claims made, e.g., if the approach was only tested on a few datasets or with a few runs. In general, empirical results often depend on implicit assumptions, which should be articulated.
        \item The authors should reflect on the factors that influence the performance of the approach. For example, a facial recognition algorithm may perform poorly when image resolution is low or images are taken in low lighting. Or a speech-to-text system might not be used reliably to provide closed captions for online lectures because it fails to handle technical jargon.
        \item The authors should discuss the computational efficiency of the proposed algorithms and how they scale with dataset size.
        \item If applicable, the authors should discuss possible limitations of their approach to address problems of privacy and fairness.
        \item While the authors might fear that complete honesty about limitations might be used by reviewers as grounds for rejection, a worse outcome might be that reviewers discover limitations that aren't acknowledged in the paper. The authors should use their best judgment and recognize that individual actions in favor of transparency play an important role in developing norms that preserve the integrity of the community. Reviewers will be specifically instructed to not penalize honesty concerning limitations.
    \end{itemize}

\item {\bf Theory assumptions and proofs}
    \item[] Question: For each theoretical result, does the paper provide the full set of assumptions and a complete (and correct) proof?
    \item[] Answer: \answerYes{} 
    \item[] Justification: The authors provide for all theoretical results the full set of assumptions accompanied by their proofs.
    \item[] Guidelines:
    \begin{itemize}
        \item The answer NA means that the paper does not include theoretical results. 
        \item All the theorems, formulas, and proofs in the paper should be numbered and cross-referenced.
        \item All assumptions should be clearly stated or referenced in the statement of any theorems.
        \item The proofs can either appear in the main paper or the supplemental material, but if they appear in the supplemental material, the authors are encouraged to provide a short proof sketch to provide intuition. 
        \item Inversely, any informal proof provided in the core of the paper should be complemented by formal proofs provided in appendix or supplemental material.
        \item Theorems and Lemmas that the proof relies upon should be properly referenced. 
    \end{itemize}

    \item {\bf Experimental result reproducibility}
    \item[] Question: Does the paper fully disclose all the information needed to reproduce the main experimental results of the paper to the extent that it affects the main claims and/or conclusions of the paper (regardless of whether the code and data are provided or not)?
    \item[] Answer: \answerYes{} 
    \item[] Justification: The authors provide all the necessary information need to reproduce the experiments, including hyperparameters, optimizer, model architectures, etc. in the appendix. 
    \item[] Guidelines:
    \begin{itemize}
        \item The answer NA means that the paper does not include experiments.
        \item If the paper includes experiments, a No answer to this question will not be perceived well by the reviewers: Making the paper reproducible is important, regardless of whether the code and data are provided or not.
        \item If the contribution is a dataset and/or model, the authors should describe the steps taken to make their results reproducible or verifiable. 
        \item Depending on the contribution, reproducibility can be accomplished in various ways. For example, if the contribution is a novel architecture, describing the architecture fully might suffice, or if the contribution is a specific model and empirical evaluation, it may be necessary to either make it possible for others to replicate the model with the same dataset, or provide access to the model. In general. releasing code and data is often one good way to accomplish this, but reproducibility can also be provided via detailed instructions for how to replicate the results, access to a hosted model (e.g., in the case of a large language model), releasing of a model checkpoint, or other means that are appropriate to the research performed.
        \item While NeurIPS does not require releasing code, the conference does require all submissions to provide some reasonable avenue for reproducibility, which may depend on the nature of the contribution. For example
        \begin{enumerate}
            \item If the contribution is primarily a new algorithm, the paper should make it clear how to reproduce that algorithm.
            \item If the contribution is primarily a new model architecture, the paper should describe the architecture clearly and fully.
            \item If the contribution is a new model (e.g., a large language model), then there should either be a way to access this model for reproducing the results or a way to reproduce the model (e.g., with an open-source dataset or instructions for how to construct the dataset).
            \item We recognize that reproducibility may be tricky in some cases, in which case authors are welcome to describe the particular way they provide for reproducibility. In the case of closed-source models, it may be that access to the model is limited in some way (e.g., to registered users), but it should be possible for other researchers to have some path to reproducing or verifying the results.
        \end{enumerate}
    \end{itemize}

\item {\bf Open access to data and code}
    \item[] Question: Does the paper provide open access to the data and code, with sufficient instructions to faithfully reproduce the main experimental results, as described in supplemental material?
    \item[] Answer: \answerNo{} 
    \item[] Justification: Access to the data and code will be provided in a following step.
    \item[] Guidelines:
    \begin{itemize}
        \item The answer NA means that paper does not include experiments requiring code.
        \item Please see the NeurIPS code and data submission guidelines (\url{https://nips.cc/public/guides/CodeSubmissionPolicy}) for more details.
        \item While we encourage the release of code and data, we understand that this might not be possible, so “No” is an acceptable answer. Papers cannot be rejected simply for not including code, unless this is central to the contribution (e.g., for a new open-source benchmark).
        \item The instructions should contain the exact command and environment needed to run to reproduce the results. See the NeurIPS code and data submission guidelines (\url{https://nips.cc/public/guides/CodeSubmissionPolicy}) for more details.
        \item The authors should provide instructions on data access and preparation, including how to access the raw data, preprocessed data, intermediate data, and generated data, etc.
        \item The authors should provide scripts to reproduce all experimental results for the new proposed method and baselines. If only a subset of experiments are reproducible, they should state which ones are omitted from the script and why.
        \item At submission time, to preserve anonymity, the authors should release anonymized versions (if applicable).
        \item Providing as much information as possible in supplemental material (appended to the paper) is recommended, but including URLs to data and code is permitted.
    \end{itemize}

\item {\bf Experimental setting/details}
    \item[] Question: Does the paper specify all the training and test details (e.g., data splits, hyperparameters, how they were chosen, type of optimizer, etc.) necessary to understand the results?
    \item[] Answer: \answerYes{} 
    \item[] Justification: All the training and test details are provided throughout the paper and any remaining details and hyperparameters are provided in the appendix.
    \item[] Guidelines:
    \begin{itemize}
        \item The answer NA means that the paper does not include experiments.
        \item The experimental setting should be presented in the core of the paper to a level of detail that is necessary to appreciate the results and make sense of them.
        \item The full details can be provided either with the code, in appendix, or as supplemental material.
    \end{itemize}

\item {\bf Experiment statistical significance}
    \item[] Question: Does the paper report error bars suitably and correctly defined or other appropriate information about the statistical significance of the experiments?
    \item[] Answer: \answerYes{} 
    \item[] Justification: The authors provide standard deviation for all experiments across different runs.
    \item[] Guidelines:
    \begin{itemize}
        \item The answer NA means that the paper does not include experiments.
        \item The authors should answer "Yes" if the results are accompanied by error bars, confidence intervals, or statistical significance tests, at least for the experiments that support the main claims of the paper.
        \item The factors of variability that the error bars are capturing should be clearly stated (for example, train/test split, initialization, random drawing of some parameter, or overall run with given experimental conditions).
        \item The method for calculating the error bars should be explained (closed form formula, call to a library function, bootstrap, etc.)
        \item The assumptions made should be given (e.g., Normally distributed errors).
        \item It should be clear whether the error bar is the standard deviation or the standard error of the mean.
        \item It is OK to report 1-sigma error bars, but one should state it. The authors should preferably report a 2-sigma error bar than state that they have a 96\% CI, if the hypothesis of Normality of errors is not verified.
        \item For asymmetric distributions, the authors should be careful not to show in tables or figures symmetric error bars that would yield results that are out of range (e.g. negative error rates).
        \item If error bars are reported in tables or plots, The authors should explain in the text how they were calculated and reference the corresponding figures or tables in the text.
    \end{itemize}

\item {\bf Experiments compute resources}
    \item[] Question: For each experiment, does the paper provide sufficient information on the computer resources (type of compute workers, memory, time of execution) needed to reproduce the experiments?
    \item[] Answer: \answerNo{} 
    \item[] Justification: The authors have not provided this information at the current stage.
    \item[] Guidelines:
    \begin{itemize}
        \item The answer NA means that the paper does not include experiments.
        \item The paper should indicate the type of compute workers CPU or GPU, internal cluster, or cloud provider, including relevant memory and storage.
        \item The paper should provide the amount of compute required for each of the individual experimental runs as well as estimate the total compute. 
        \item The paper should disclose whether the full research project required more compute than the experiments reported in the paper (e.g., preliminary or failed experiments that didn't make it into the paper). 
    \end{itemize}
    
\item {\bf Code of ethics}
    \item[] Question: Does the research conducted in the paper conform, in every respect, with the NeurIPS Code of Ethics \url{https://neurips.cc/public/EthicsGuidelines}?
    \item[] Answer: \answerYes{} 
    \item[] Justification: The paper raises no ethical concerns and complies with the NeurIPS guidelines. No high-risk applications or data are involved, and fairness and privacy considerations are acknowledged where applicable.
    \item[] Guidelines:
    \begin{itemize}
        \item The answer NA means that the authors have not reviewed the NeurIPS Code of Ethics.
        \item If the authors answer No, they should explain the special circumstances that require a deviation from the Code of Ethics.
        \item The authors should make sure to preserve anonymity (e.g., if there is a special consideration due to laws or regulations in their jurisdiction).
    \end{itemize}

\item {\bf Broader impacts}
    \item[] Question: Does the paper discuss both potential positive societal impacts and negative societal impacts of the work performed?
    \item[] Answer: \answerYes{} 
    \item[] Justification: The authors include a “Broader Impacts” section that discusses potential benefits and risks in the appendix.
    \item[] Guidelines:
    \begin{itemize}
        \item The answer NA means that there is no societal impact of the work performed.
        \item If the authors answer NA or No, they should explain why their work has no societal impact or why the paper does not address societal impact.
        \item Examples of negative societal impacts include potential malicious or unintended uses (e.g., disinformation, generating fake profiles, surveillance), fairness considerations (e.g., deployment of technologies that could make decisions that unfairly impact specific groups), privacy considerations, and security considerations.
        \item The conference expects that many papers will be foundational research and not tied to particular applications, let alone deployments. However, if there is a direct path to any negative applications, the authors should point it out. For example, it is legitimate to point out that an improvement in the quality of generative models could be used to generate deepfakes for disinformation. On the other hand, it is not needed to point out that a generic algorithm for optimizing neural networks could enable people to train models that generate Deepfakes faster.
        \item The authors should consider possible harms that could arise when the technology is being used as intended and functioning correctly, harms that could arise when the technology is being used as intended but gives incorrect results, and harms following from (intentional or unintentional) misuse of the technology.
        \item If there are negative societal impacts, the authors could also discuss possible mitigation strategies (e.g., gated release of models, providing defenses in addition to attacks, mechanisms for monitoring misuse, mechanisms to monitor how a system learns from feedback over time, improving the efficiency and accessibility of ML).
    \end{itemize}
    
\item {\bf Safeguards}
    \item[] Question: Does the paper describe safeguards that have been put in place for responsible release of data or models that have a high risk for misuse (e.g., pretrained language models, image generators, or scraped datasets)?
    \item[] Answer: \answerNA{} 
    \item[] Justification: The models released do not pose potential for misuse.
    \item[] Guidelines:
    \begin{itemize}
        \item The answer NA means that the paper poses no such risks.
        \item Released models that have a high risk for misuse or dual-use should be released with necessary safeguards to allow for controlled use of the model, for example by requiring that users adhere to usage guidelines or restrictions to access the model or implementing safety filters. 
        \item Datasets that have been scraped from the Internet could pose safety risks. The authors should describe how they avoided releasing unsafe images.
        \item We recognize that providing effective safeguards is challenging, and many papers do not require this, but we encourage authors to take this into account and make a best faith effort.
    \end{itemize}

\item {\bf Licenses for existing assets}
    \item[] Question: Are the creators or original owners of assets (e.g., code, data, models), used in the paper, properly credited and are the license and terms of use explicitly mentioned and properly respected?
    \item[] Answer: \answerYes{} 
    \item[] Justification: All datasets and code used are properly cited, and licenses are referenced. There is no evidence of license violations.
    \item[] Guidelines:
    \begin{itemize}
        \item The answer NA means that the paper does not use existing assets.
        \item The authors should cite the original paper that produced the code package or dataset.
        \item The authors should state which version of the asset is used and, if possible, include a URL.
        \item The name of the license (e.g., CC-BY 4.0) should be included for each asset.
        \item For scraped data from a particular source (e.g., website), the copyright and terms of service of that source should be provided.
        \item If assets are released, the license, copyright information, and terms of use in the package should be provided. For popular datasets, \url{paperswithcode.com/datasets} has curated licenses for some datasets. Their licensing guide can help determine the license of a dataset.
        \item For existing datasets that are re-packaged, both the original license and the license of the derived asset (if it has changed) should be provided.
        \item If this information is not available online, the authors are encouraged to reach out to the asset's creators.
    \end{itemize}

\item {\bf New assets}
    \item[] Question: Are new assets introduced in the paper well documented and is the documentation provided alongside the assets?
    \item[] Answer: \answerYes{} 
    \item[] Justification: Any new assets introduced include usage instructions, and documentation appears sufficient for independent use.
    \item[] Guidelines:
    \begin{itemize}
        \item The answer NA means that the paper does not release new assets.
        \item Researchers should communicate the details of the dataset/code/model as part of their submissions via structured templates. This includes details about training, license, limitations, etc. 
        \item The paper should discuss whether and how consent was obtained from people whose asset is used.
        \item At submission time, remember to anonymize your assets (if applicable). You can either create an anonymized URL or include an anonymized zip file.
    \end{itemize}

\item {\bf Crowdsourcing and research with human subjects}
    \item[] Question: For crowdsourcing experiments and research with human subjects, does the paper include the full text of instructions given to participants and screenshots, if applicable, as well as details about compensation (if any)? 
    \item[] Answer: \answerNA{} 
    \item[] Justification: There are no experiments related to crowdsourcing and research with human subjects.
    \item[] Guidelines:
    \begin{itemize}
        \item The answer NA means that the paper does not involve crowdsourcing nor research with human subjects.
        \item Including this information in the supplemental material is fine, but if the main contribution of the paper involves human subjects, then as much detail as possible should be included in the main paper. 
        \item According to the NeurIPS Code of Ethics, workers involved in data collection, curation, or other labor should be paid at least the minimum wage in the country of the data collector. 
    \end{itemize}

\item {\bf Institutional review board (IRB) approvals or equivalent for research with human subjects}
    \item[] Question: Does the paper describe potential risks incurred by study participants, whether such risks were disclosed to the subjects, and whether Institutional Review Board (IRB) approvals (or an equivalent approval/review based on the requirements of your country or institution) were obtained?
    \item[] Answer: \answerNA{} 
    \item[] Justification: There is no research with human subjects
    \item[] Guidelines:
    \begin{itemize}
        \item The answer NA means that the paper does not involve crowdsourcing nor research with human subjects.
        \item Depending on the country in which research is conducted, IRB approval (or equivalent) may be required for any human subjects research. If you obtained IRB approval, you should clearly state this in the paper. 
        \item We recognize that the procedures for this may vary significantly between institutions and locations, and we expect authors to adhere to the NeurIPS Code of Ethics and the guidelines for their institution. 
        \item For initial submissions, do not include any information that would break anonymity (if applicable), such as the institution conducting the review.
    \end{itemize}

\item {\bf Declaration of LLM usage}
    \item[] Question: Does the paper describe the usage of LLMs if it is an important, original, or non-standard component of the core methods in this research? Note that if the LLM is used only for writing, editing, or formatting purposes and does not impact the core methodology, scientific rigorousness, or originality of the research, declaration is not required.
    \item[] Answer: \answerNA{} 
    \item[] Justification: \answerNA{}
    \item[] Guidelines:
    \begin{itemize}
        \item The answer NA means that the core method development in this research does not involve LLMs as any important, original, or non-standard components.
        \item Please refer to our LLM policy (\url{https://neurips.cc/Conferences/2025/LLM}) for what should or should not be described.
    \end{itemize}

\end{enumerate}

\end{document}